\documentclass{article}

\usepackage{PRIMEarxiv}

\usepackage[utf8]{inputenc} 
\usepackage[T1]{fontenc}    
\usepackage{hyperref}       
\usepackage{url}            
\usepackage{booktabs}       
\usepackage{amsfonts}       
\usepackage{nicefrac}       
\usepackage{microtype}      
\usepackage{lipsum}
\usepackage{fancyhdr}       
\usepackage{graphicx}       
\graphicspath{{media/}}     

\usepackage{graphicx}%
\usepackage{multirow}%
\usepackage{amsmath,amssymb,amsfonts}%
\usepackage{amsthm}%
\usepackage{bbm}
\usepackage{mathrsfs}%
\usepackage[title]{appendix}%
\usepackage{xcolor}%
\usepackage{textcomp}%
\usepackage{manyfoot}%
\usepackage{booktabs}%
\usepackage{algorithm}%
\usepackage{algorithmicx}%
\usepackage{algpseudocode}%
\usepackage{listings}%
\usepackage{bm}
\usepackage{tablefootnote}

\newcommand{\ub}[1]{\underline{\textbf{#1}}}
\newcommand{\ux}[1]{\underline{#1}}

\newcommand{\update}[1]{{#1}}

\title{Analytical Softmax Temperature Setting from Feature Dimensions for Model- and Domain-Robust Classification
\thanks{\textit{\underline{Citation}}: 
\textbf{T. Hasegawa, S. Sakai, "Analytical Softmax Temperature Setting from Feature Dimensions for Model- and Domain-Robust Classification", xxx under review. pp. x--x, DOI:000000/11111.}} 
}

\author{
  Tatsuhito Hasegawa \\
  Graduate School of Engineering, \\
  University of Fukui, \\
  Fukui\\
  \texttt{t-hase@u-fukui.ac.jp} \\
   \And
  Shunsuke Sakai \\
  Graduate School of Engineering, \\
  University of Fukui, \\
  Fukui\\
  \texttt{mf240599@g.u-fukui.ac.jp} \\
}

\begin{document}
\maketitle

\begin{abstract}
In deep learning-based classification tasks, the softmax function’s temperature parameter $T$ critically influences the output distribution and overall performance. This study presents a novel theoretical insight that the optimal temperature $T^*$ is uniquely determined by the dimensionality of the feature representations, thereby enabling training-free determination of $T^*$. Despite this theoretical grounding, empirical evidence reveals that $T^*$ fluctuates under practical conditions owing to variations in models, datasets, and other confounding factors. To address these influences, we propose and optimize a set of temperature determination coefficients that specify how $T^*$ should be adjusted based on the theoretical relationship to feature dimensionality. Additionally, we insert a batch normalization layer immediately before the output layer, effectively stabilizing the feature space. Building on these coefficients and a suite of large-scale experiments, we develop an empirical formula to estimate $T^*$ without additional training while also introducing a corrective scheme to refine $T^*$ based on the number of classes and task complexity. Our findings confirm that the derived temperature not only aligns with the proposed theoretical perspective but also generalizes effectively across diverse tasks, consistently enhancing classification performance and offering a practical, training-free solution for determining $T^*$.
\end{abstract}

\keywords{Temperature Determination Coefficients, Optimal Temperature Parameter, Softmax Function, Training-free hyperparameter selection}

\section{Introduction}
\begin{figure}[h]
\centering
 \includegraphics[width=.75\textwidth]{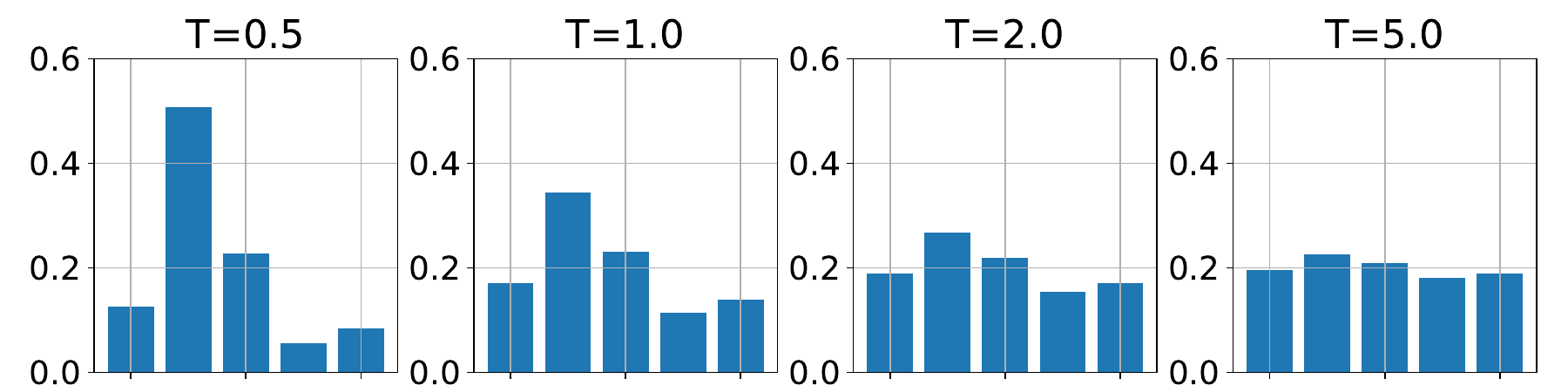}
 \caption{Sample outputs of the softmax function at various temperatures for a fixed model output $\hat{{\bm y}}$.}
 \label{fig:softmax}
\end{figure}

Deep learning exhibits impressive performance in various areas, such as image recognition \cite{kriz}, object detection \cite{liu}, image generation \cite{karras}, and natural language processing \cite{NEURIPS2020_1457c0d6}.
In general, a deep learning model can be constructed by stacking layers, such as a convolutional layer, a fully connected layer, and a normalization layer. This model is then trained by minimizing the expected risk using a defined loss function. The cross-entropy loss displayed in Eq. \eqref{eq:ce} is typically used for standard classification:
\begin{equation}
\label{eq:ce}
 \mathcal{L}(f({\bm x};{\bm \theta}),{\bm y})=-\frac{1}{n}\sum_{i=1}^{n}\sum_{j=1}^{c}y_{ij}\log f_j({\bm x}_i;{\bm \theta}),
\end{equation}
where $n$ and $c$ represent the number of samples and categories, respectively;
$\bm \theta$ denotes the parameters of the deep neural network; $f_j({\bm x}_i;{\bm \theta})$ denotes the $j$-th output of the model corresponding to input $\bm x_i$; and $y_{ij} \in \{0, 1\}$ indicates whether the $i$-th sample belongs to category $j$. Because the cross-entropy loss takes a true one-hot distribution $\bm y$ and an estimated distribution, the output of the model $\hat{{\bm y}}$ is normalized by the softmax function, as expressed in Eq. \eqref{eq:softmax}:

\begin{equation}
\label{eq:softmax}
 \operatorname{softmax}(\hat{y}_j,\hat{{\bm y}})=\frac{\exp(\frac{\hat{y}_j}{T})}{\sum_{\hat{y}_k\in  \hat{{\bm y}}}\exp(\frac{\hat{y}_k}{T})},
\end{equation}
where $T$ is a hyperparameter of the softmax function, known as the temperature. In Fig. \ref{fig:softmax}, we illustrate the effect of different values of $T$ when the model's output $\hat{{\bm y}}$ is fixed. Intuitively, the temperature $T$ controls the uncertainty of the estimated distribution. As $T$ increases, the estimated distribution approaches a uniform distribution; conversely, as $T$ decreases, the estimated distribution becomes deterministic. In many settings, the temperature $T$ is set to a default value of 1.0. 
Examples of applications involving $T$ include knowledge distillation \cite{kd}, which transfers knowledge from a larger model to a smaller model, and temperature scaling \cite{ts}, which adjusts the model's confidence. While some applications control the temperature to meet specific requirements, the effect of temperature on the model's learning process remains poorly understood. 
\update{However, the temperature parameter $T$ used during the training process is a crucial factor that significantly impacts the model's final generalization performance. T adjusts the sharpness of the probability distribution output by the softmax function, which effectively scales the gradients calculated through the cross-entropy loss. This gradient scaling directly influences the learning dynamics, convergence stability, and ultimately, how well the model performs on unseen data (i.e., generalization performance). Furthermore, an appropriate value of $T$ can act as a form of regularization, potentially preventing the model from overfitting to the training data and helping it learn more generalizable representations. Therefore, \textbf{selecting a suitable $T$ for the task and model is essential not only for adjusting the model's output distribution but also for optimizing the training process itself to build models with better generalization capabilities}. Particularly in many practical scenarios where high generalization performance on unseen data is critically important, such as large-scale image classification \cite{kriz}, natural language understanding \cite{NEURIPS2020_1457c0d6}, or time-series forecasting in changing environments \cite{Ismail2019}, optimizing the temperature parameter during training is crucial for enhancing model reliability and utility.}

Some studies \cite{agarwala2023temperature, hase} investigated the effect of changes in temperature parameters on model training. Agarwala \emph{et~al.}~\cite{agarwala2023temperature} tackled image classification tasks by introducing the inverse temperature $\beta$ as a hyperparameter and, through extensive experimentation on various tasks, demonstrated that tuning $\beta$ can significantly improve model performance. While they established the importance of $\beta$ for optimizing generalization, no universal formula or method was put forward to pinpoint the best $\beta$ across disparate model architectures and datasets. Concurrently, Hasegawa 
\cite{hase} explored a possible correlation between the optimal temperature $T$ and the dimensionality $M$ of the feature map in a 1D CNN-based human activity recognition scenario and showed a trend in which $T$ grows with $M$, suggesting that these two factors may be functionally linked. However, variations in tasks and model architectures introduced inconsistencies, rendering it difficult to derive a single, general rule for determining $T$. Taken together, these findings underscore both the necessity and the complexity of fine-tuning the temperature parameter. Despite these advances, the community still lacks a systematic, generalizable method to determine either $T$ or $\beta$ robustly in the face of architectural and task-related differences. 

In this study, we propose a novel method for estimating the optimal temperature parameter $T^*$ in a \textit{training-free} manner that is robust to variations in both model architectures and tasks. By eliminating the need for extensive tuning, our approach not only reduces the computational burden but also achieves superior performance compared to the commonly used default $T=1$. Because minimizing hyperparameters is critical for the practical deployment of deep learning models, this study represents a significant step toward more efficient and scalable deep learning solutions. \update{Since the optimal value of the temperature \(T\), a hyperparameter directly involved in the training process, typically depends heavily on the dataset and model architecture \cite{agarwala2023temperature,xuan2025exploringimpacttemperaturescaling}, tuning it conventionally requires computationally expensive trial‑and‑error with multiple full model training runs. Our method is the first to provide a training‑free, closed‑form estimator for the optimal temperature \(T^*\), leveraging feature dimensionality and task characteristics. By eliminating costly searches, it offers significant practical value—substantially saving computational resources and development time—while also introducing a principled theoretical framework for temperature determination in deep learning.}

The main contributions of this study are as follows:

\begin{itemize}
    \item \textbf{A theoretical framework linking $T^*$ and feature dimensionality.}  
    We establish a theoretical relationship between the optimal temperature parameter $T^*$ and the feature dimensionality $M$. In addition, we propose to represent $T^*$ using a set of \emph{temperature determination coefficients} $\alpha, \beta, \gamma, \delta$ as
    \[
    T^* = \alpha \sqrt{M} \;+\; \beta \;+\; \gamma \,\log(\textit{csg}) \;+\; \delta \,\log(\textit{cn}),
    \]
    where $\alpha, \beta, \gamma, \delta$ are experimentally optimized constants, and $\textit{csg}$, $\textit{cn}$ denote the cumulative spectral gradient (CSG)~\cite{Frederic2019} and the number of classes (CN), respectively. Furthermore, we show that inserting a batch normalization (BN) layer \cite{bn} immediately before the output layer effectively stabilizes this relationship, making $T^*$ estimation less sensitive to changes in model architecture and dataset characteristics.
    
    \item \textbf{Empirical analysis of BN insertion and its performance impact.}  
    Through extensive experiments, we demonstrate that while placing a BN layer right before the output layer may cause performance degradation under the conventional setting \(T=1\), it can improve performance when the temperature \(T^*\) is set appropriately. This finding emphasizes BN’s dual role of normalizing the feature space and facilitating more precise temperature selection, which in turn leads to improved generalization.
    
    \item \textbf{An empirical formula for training-free estimation of \(T^*\).}  
    By leveraging large-scale experiments, we optimize the \emph{temperature determination coefficients} \(\alpha, \beta\) to propose a closed-form solution for \(\hat{T}^*\) that requires no additional training overhead:
    \[
    \hat{T}^*  = \operatorname{clip}\bigl(0.7239 \,\sqrt{M} - 4.706,\; \epsilon,\; 512\bigr),
    \]
    which consistently outperforms the conventional default \(T=1\). This formula serves as a reliable starting point for determining \(T^*\), eliminating the need for time-consuming hyperparameter searches.
    
    \item \textbf{Task-aware refinement using temperature determination coefficients.}  
    We further refine \(\hat{T}^*\) by incorporating two task-dependent terms: the number of classes (\(cn\)) and the cumulative spectral gradient (\(csg\)). Specifically, we again optimize the temperature determination coefficients \(\alpha, \beta, \gamma, \delta\) using large-scale experimental results to obtain:
    \[
    \hat{T}^*_{csgcn} = \operatorname{clip}\Bigl(0.3191 \,\sqrt{M} + 20.74 + 3.746 \,\log(csg) - 7.380 \,\log(cn),\; \epsilon,\; 512\Bigr).
    \]
    By accounting for both task complexity and the number of classes, this refined scheme offers robust and generalizable performance gains across a wide range of deep learning applications.
\end{itemize}

\section{Related Works}
\subsection{Temperature parameter}
\label{sec:prev}
Several studies have utilized the temperature of the softmax function, with knowledge distillation (KD) \cite{kd, liu2022metaKD, curriculum2023, sun2024logit} and temperature scaling \cite{ts, Balanya2024} serving as representative examples. Knowledge distillation is a method for transferring knowledge from a teacher model to a student model. This method aims to improve performance and reduce the number of model parameters. \update{In conventional KD, it is well known that setting the softmax temperature to approximately \(T=4\) during distillation improves performance. Liu \emph{et al.}~\cite{liu2022metaKD} proposed Meta KD, which optimizes \(T\) via meta‑gradients computed on a validation set. Curriculum Temperature KD (CTKD) \cite{curriculum2023} introduces an adversarial learning scheme that directly learns the temperature parameter itself. Sun \emph{et al.}~\cite{sun2024logit} achieve a dynamic temperature by applying logit standardization independently to both the teacher and the student models. }
Furthermore, temperature scaling \cite{ts} reduces the model's prediction bias by correcting temperature parameters. \update{Balanya \textit{et al.} \cite{Balanya2024} utilized adaptive temperature in the temperature scaling. In contrastive learning, the temperature parameter is also adjusted to modulate the influence of negative samples \cite{Wang2021CL}.
}

In addition to these approaches, Agarwala \emph{et~al.}~\cite{agarwala2023temperature} presented a temperature check method examining the effect of the inverse temperature $\beta=1/T$ on model generalization, highlighting the importance of temperature adjustment. The authors tuned the inverse temperature during standard model training within the range $\beta \in [10^{-2}, 10^{1}]$. \update{Xuan \emph{et~al.} \cite{xuan2025exploringimpacttemperaturescaling} also highlighted the importance of tuning the temperature parameter in classification tasks.} Furthermore, relaxed softmax \cite{Neumann2018-dk} also proposes a tuning strategy for inverse temperature. In fields other than classification, temperature parameters have attracted attention in deep metric learning. Xu \cite{Zhang2018-rp} proposed a heated-up strategy that involves training the model with increasing temperature. 

However, none of the above studies discusses the relationship between the dimensionality of the feature maps $M$ and the optimal temperature $T^*$. While these studies note the importance of tuning the temperature parameter, they do not sufficiently address the question of whether the optimal temperature parameter can be estimated without training the model.

Building upon these general insights into temperature tuning, 
a study in sensor-based human activity recognition has explored the relationship between feature dimensionality and temperature \cite{hase}. Denoting the number of dimensions of the output of the feature extractor (e.g., ConvNets) as $M$, the experimental results indicate a potential relationship between the optimal temperature $T^*$ and $M$. However, there were several limitations: 
\begin{enumerate}
\item{
The experiments were conducted only on one-dimensional sensor-based human activity recognition datasets. \label{lim1}
} 
\item{
The coefficients determining the optimal temperature parameter vary across different datasets and the encoder's model architectures. \label{lim2}
}
\item{
The effectiveness has not been evaluated using a deeper model. \label{lim3}
}
\end{enumerate}

For these limitations, 
the related study \cite{hase} discusses a method for mitigating limitation (\ref{lim2}) by inserting layer normalization (LN) \cite{ba2016layer} before applying the softmax function. This method is inspired by the observation that an output distribution with optimal temperature $T^*$ follows a specific distribution. By normalizing the outputs and dynamically adjusting the temperature parameter through the trainable parameters in LN, this method can be regarded as an advanced extension of the relaxed softmax approach \cite{Neumann2018-dk}. However, it does not account for the potential relationship between $M$ and $T^*$. In this paper, we further explore the relationship between $M$ and $T^*$. 

\subsection{Scaled dot-product attention}
The scaled dot-product attention, introduced by Vaswani \textit{et al.} \cite{Vaswani2017}, focuses on the temperature parameter and the number of input dimensions of the softmax function. This operation can be computed as follows: 
\begin{equation}
    \operatorname{Attention}(Q, K, V) = \operatorname{softmax}(\frac{QK^T}{\sqrt{d_k}})V,
\end{equation}
where $Q$ and $K^T$ denote the feature map ${\bm z}$ and parameters ${\bm w_j}$, respectively. Here, the problem of increasing dispersion occurs as the number of dimensions increases, as discuss in the next section. The authors addressed this problem by dividing by $\sqrt{d_k}$, the square root of the number of dimensions of $Q$ and $K$. This is equivalent to the softmax function with $T = \sqrt{d_k}$. The authors introduced this temperature adjusting to maintain an appropriate gradient scale. However, they did not provide a detailed discussion of this temperature adjustment. 

\subsection{Label smoothing}
\label{sec:smooth}
Label smoothing \cite{Szegedy2016} is a regularization method that has a role similar to that of temperature scaling. As displayed in Eq. \eqref{eq:ce}, a typical cross-entropy loss function employs ground-truth labels $\boldsymbol{y} \in \{0, 1\}^{c}$, which are encoded in a one-hot manner. In contrast, label smoothing represents ground-truth labels $\boldsymbol{y}^{LS} \in \mathbb{R}^{c}$ in a soft manner, as expressed in Eq. \eqref{eq:ls}:
\begin{equation}
    y_i^{LS} = y_i(1-\epsilon) + \epsilon/c.
\label{eq:ls}
\end{equation}
Label smoothing minimizes interclass variance, maximizes intraclass variance, and improves the model's generalization ability \cite{Rafael2019}. In addition to a method for statically constructing soft labels from a uniform distribution, a dynamic label construction method has also been proposed \cite{Zhang2021}. 

\begin{figure}[h]
 \centering
 \includegraphics[width=0.7\textwidth]{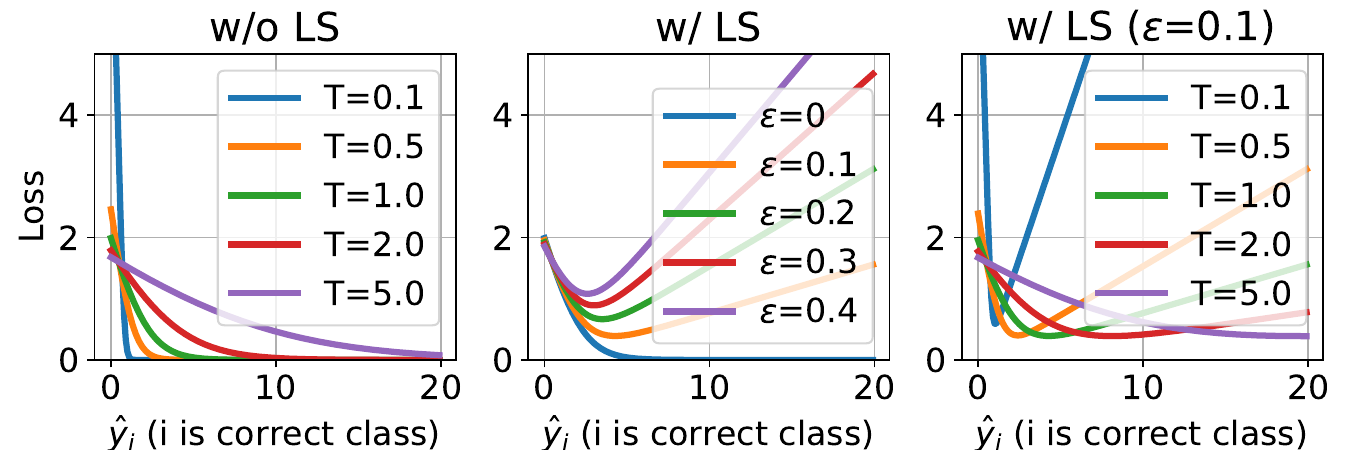}
 \caption{Effect of temperature variation and label smoothing on cross-entropy loss.}
 \label{fig:labelsmoothing}
\end{figure}

Temperature adjusting and label smoothing are similar in that they focus on the distribution of negative samples. However, whereas temperature adjusting controls the estimated distribution $\hat{\boldsymbol{y}}$, label smoothing transforms the ground-truth distribution $\boldsymbol{y}$. In Fig. \ref{fig:labelsmoothing}, we visualize a simulation of the response of the cross-entropy loss as the logits $\hat{y}_i$ are varied. In the case without label smoothing, we observe that the loss monotonically decreases, with temperature regulating extreme changes in the loss. In the case of label smoothing, we observe that label smoothing penalizes large logits $\hat{y}_i$ in response to $\epsilon$. As illustrated in the figure, the effects of both regularizations are observed in the case involving label smoothing ($\epsilon = 0.1$). 

Based on the above observations, we conclude that label smoothing and temperature adjusting serve distinct roles in model training. Specifically, we focus on the relationship between the dimensionality of the feature map $M$ and the optimal temperature $ T^*$. 

\subsection{Position of this study}
In this section, we clarify the position of our study. Although knowledge distillation~\cite{kd} and temperature scaling~\cite{ts} have demonstrated the effectiveness of tuning the temperature parameter, both methods primarily focus on distillation and calibration rather than classification. Scaled dot-product attention~\cite{Vaswani2017}, on the other hand, replaces the temperature with the square root of the feature dimensionality $d_k$ in the softmax function. However, it is treated as part of the model architecture and does not involve deriving an explicit optimal temperature for classification. Furthermore, although relaxed softmax~\cite{Neumann2018-dk} and inserting LN~\cite{hase} dynamically adjusts the temperature parameter during training, a theoretical framework linking the feature dimensionality $M$ to the optimal temperature $T^*$ has not been established. \update{As such, existing research has not theoretically established the optimal temperature \(T^*\) for improving generalization performance in classification tasks.}

To address this gap, we propose a new approach that does not require additional training and incorporates the effect of $M$ into a closed-form estimation of the optimal temperature. As summarized in Table~\ref{table:characteristics}, this study formalizes and exploits the previously underexplored relationship between feature dimensionality and the temperature parameter, setting it apart from conventional methods. Our goal is to achieve consistent performance improvements across various model architectures and application tasks by offering a theoretically grounded and practically efficient temperature estimation procedure. \update{In particular, while temperature check~\cite{agarwala2023temperature} determines \(T^*\) through costly hyperparameter searches involving extensive model retraining, our proposed method is novel in that it uniquely identifies \(T^*\) in a training‑free manner. Although Hasegawa \textit{et al.} \cite{hase} also noted a relationship between \(M\) and \(T\), they did not derive a closed‑form solution. By contrast, our approach defines an explicit closed-form solution and further incorporates corrections for task difficulty and class count, representing a novel contribution.}

\begin{table}[h] 
\centering
    \caption{Characteristics of our method compared with related works.}
    \label{table:characteristics}
    \begin{tabular}{@{}l|cccc}
    \toprule
\textbf{Method} &	\textbf{Task} &	\textbf{Tuning} &	\textbf{Consider $M$} &	\textbf{Closed-form} \\ \midrule
KD \cite{kd} &	Distillation &	Required &	No &	No \\ 
\update{Meta KD \cite{liu2022metaKD}} &	Distillation &	Dynamic &	No &	No \\ 
\update{CTKD \cite{curriculum2023}} &	Distillation &	Dynamic &	No &	No \\ 
\update{Logit standardization \cite{sun2024logit}} &	Distillation &	Dynamic &	No &	No \\ 
Temperature Scaling \cite{ts} &	Calibration &	Required &	No &	No \\ 
Temperature Check \cite{agarwala2023temperature, xuan2025exploringimpacttemperaturescaling} &	Classification &	Required &	No &	No \\ 
Relaxed Softmax \cite{Neumann2018-dk} &	Classification &	Dynamic &	No &	No \\ 
Scaled Dot-Product Attn \cite{Vaswani2017} &	Architecture &	Unnecessary &	Yes &	Yes \\ 
Label Smoothing \cite{Szegedy2016} &	Classification &	Required &	No &	No \\ 
Insert LN \cite{hase} &	Classification &	Dynamic &	No &	No \\ 
Ours &	Classification &	\textbf{Unnecessary} &	\textbf{Yes} &	\textbf{Yes} \\ 
        \bottomrule
    \end{tabular}
\end{table}

\section{Training-free temperature determination}
\update{As demonstrated in related works, tuning the temperature parameter of the softmax function is critical for improving performance in classification tasks \cite{agarwala2023temperature, hase}; however, its optimal value has traditionally been chosen empirically---either fixed at 1 or selected via exploratory methods such as grid search. In this study, we introduce a training‑free temperature determination approach that identifies the optimal temperature without any hyperparameter search. Our method is exceptionally easy to integrate, requires no modifications to existing training pipelines, and is broadly applicable to any deep learning model employing softmax cross‑entropy loss. Moreover, it rests on a transparent theoretical framework, making its behavior both predictable and interpretable.}

\update{Implementing our approach entails only the following simple steps:}

\begin{figure}[h]
\centering
 \includegraphics[width=0.5\textwidth]
      {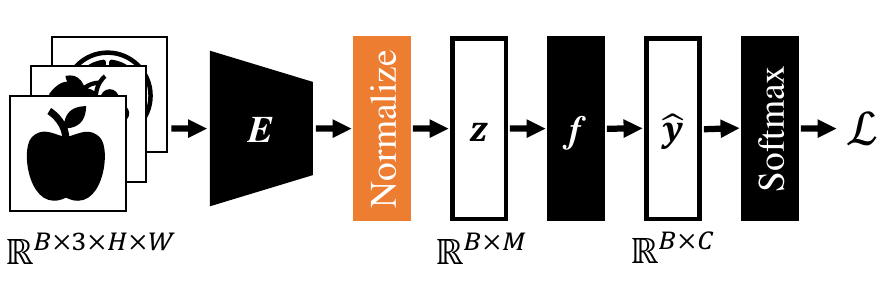}
 \caption{Outline of the flow of the general deep neural network models inserted a normalization layer.}
 \label{fig:model}
\end{figure}

\begin{enumerate}
    \item \update{Inserting BN just before the output layer as illustrated in Fig. \ref{fig:model}.}
    \item \update{Determining the temperature by following equation:}
    \begin{align}
      T^* 
      &= \underset{\color{blue}\text{Model correction}}%
                 {\underline{\alpha\,\sqrt{M}}}
       + \underset{\color{red}\text{Bias term}}%
                 {\underline{\beta}}
       + \underset{\color{orange}\text{Task correction}}%
                 {\underline{\gamma\,\log(\mathrm{csg}) + \delta\,\log(\mathrm{cn})}},
      \label{eq:final}
    \end{align}
\end{enumerate}
where $\alpha, \beta, \gamma, \delta$ are extended-version of temperature determination coefficients, and $\textit{csg}$, $\textit{cn}$ denote the CSG \cite{Frederic2019} and the number of classes, respectively. \update{The first term compensates for changes in the feature‐map dimensionality \(M\), based on the hypothesis that \(M\) exerts a dominant influence on the determination of \(T^*\) (see Section~\ref{sec:dims}). Furthermore, under the hypothesis that this dimensionality correction alone cannot fully remove the effects of differing model architectures (e.g., ResNet, ViT, etc), we insert a batch‑normalization layer immediately before the output layer (see Section~\ref{sec:bninsert}). The second term serves as a bias to adjust the overall baseline. The third and fourth terms correct for differences in task characteristics (see Section~\ref{sec:tasks}).}

\subsection{\update{Model correction by $\alpha \sqrt{M}$}} 
\label{sec:dims}
\update{$T^*$ has been shown to depend on the feature‐map dimensionality $M$ in related work~\cite{hase}. In this section, we theoretically motivate the need to introduce the model‐correction term (the first term) in Eq.~\eqref{eq:final}.}

In general, a linear classifier computes the logits from the output 
\(\bm{z} \in \mathbb{R}^M\) of the feature extractor \(E\) as follows:
\begin{equation}
\label{eq:linear}
\hat{y}_j = \bm{W}_j \bm{z}^T + b_j,
\end{equation}
where \(\bm{W}_j = [w_{j1}, w_{j2}, \ldots, w_{jM}]\) is an \(M\)-dimensional weight vector corresponding to category \(j\). 

At a given training step, the expectation and variance of \(\hat{y}_j\) are given by:
\begin{align}
\label{eq:exp2}
    \mathbb{E}[\hat{y}_j] 
    &= \mathbb{E}\Bigl[ \sum_{k=1}^{M} w_{jk} z_k + b_j \Bigr] 
     = \sum_{k=1}^{M} w_{jk}\,\mathbb{E}[z_k] + b_j, \\
\label{eq:var2}
    \mathbb{V}[\hat{y}_j] 
    &= \mathbb{V}\Bigl[ \sum_{k=1}^{M} w_{jk} z_k + b_j \Bigr] 
     = \mathbb{V}\Bigl[ \sum_{k=1}^{M} w_{jk} z_k \Bigr].
\end{align}

Here, \(w\) is treated as a constant, and \(z\) is a random variable that is generally neither independent nor normalized. In contrast, at the initial stage of training, \(w\) and \(z\) can often be approximated as independent and identically distributed. Under this assumption, the expectation and variance can be simplified to:
\begin{align}
\label{eq:exp}
    \mathbb{E}[\hat{y}_j] &= M\,\mathbb{E}[w_j z] + b_j, \\
\label{eq:var}
    \mathbb{V}[\hat{y}_j] &= M\,\mathbb{V}[w_j z].
\end{align}

Because this variance grows proportionally to \(M\), increasing model depth can lead to exploding or vanishing gradients. To address this issue, LeCun's initialization~\cite{LeCun2012} sets the initial parameters according to \(\mathbb{V}[w] = 1/M\), an approach that was subsequently extended in the Xavier~\cite{glorot2010} and He~\cite{init} initialization methods.

Focusing on the effect at the output layer, \(\hat{y}_j\) is passed into the softmax function in Eq.~(\ref{eq:softmax}). In other words, if \(\hat{y}_j\) \textbf{depends on \(M\)}, it \textbf{implicitly modifies} the temperature parameter \(T\) as a function of \(M\). Consequently, this suggests that \(T\) should be \textbf{chosen appropriately based on \(M\)}. In this study, we introduce \textbf{temperature determination coefficients} ($\alpha$ and $\beta$) to determine the optimal temperature $T^*$ as a function of $M$ as follows:
\begin{align}
T^* = \alpha \sqrt{M} \;+\; \beta.
\end{align}
The variance of $\hat{y}_j / T^*$ within the softmax function is expressed by the following equation.
\begin{align}
\label{eq:vat_tast}
    \mathbb{V}[\hat{y}_j/T^*] 
    &= \frac{1}{(\alpha \sqrt{M} + \beta)^2}\mathbb{V}\Bigl[ \sum_{k=1}^{M} w_{jk} z_k \Bigr].
\end{align}
When $\alpha = 1.0$ and $\beta = 0.0$, the variance becomes $1/M$, thereby suppressing the influence of $M$ on the softmax function. On the other hand, while $\mathbb{V}[\hat{y}_j/T^*]$ is not affected by $M$, it cannot be guaranteed to represent the optimal distribution for training when using the softmax cross-entropy loss. Therefore, the optimal $T^*$ for training is determined by adjusting the temperature determination coefficients.

\subsection{Inseting BN just before the output layer}
\label{sec:bninsert}
\subsubsection{Implementation}
As illustrated in Fig. \ref{fig:model}, The model architecture is simple, inserting the normalization layer just before the output layer. Considering image recognition as an example, the standard model takes an input image ${\bm x} \in \mathbb{R}^{B \times 3 \times H \times W}$, extracts the feature maps ${\bm z} \in \mathbb{R}^{B \times M}$ using the feature extractor $E$, and ultimately obtains the output ${\bm y} \in \mathbb{R}^{B \times C}$ using the classifier $f$. In this case, we apply a global average pooling (GAP) layer to the output of the feature extractor. We can use any model architecture as a feature extractor, such as ConvNets and vision transformers (ViTs). For simplicity, we use a single fully connected layer for the classifier $f$. 

\subsubsection{Hypothesis}
\label{sec:hypothesis}
In 
the related study \cite{hase}, comprehensive validation experiments demonstrated that when the model architecture or task differs, $T^*$ behaves as a function of $M$, though its trend undergoes slight variations (i.e., the temperature determination coefficients fluctuate). We hypothesize that this may be due to variations in $z_k$ in Eq. (\ref{eq:vat_tast}) depending on the model or task.

In this study, we consider normalizing $\boldsymbol{z}$ to enhance the robustness of temperature determination coefficients against variations in model architecture and task. 

First, by transforming Eq. (\ref{eq:var2}), the following equation can be derived.
\begin{align}
\label{eq:var3}
\mathbb{V}\biggl[\sum_{k=1}^{M}w_{jk} z_k\biggr] & = \mathbb{E}\biggl[\biggl\{\sum_{k=1}^{M}w_{jk}z_k\biggr\}^2\biggr] - \mathbb{E}\biggl[\biggl\{\sum_{k=1}^{M}w_{jk}z_k\biggr\}\biggr]^2 \nonumber \\
 & = \sum_{k=1}^{M}\sum_{l=1}^{M}w_{jk}w_{jl}\mathbb{E}[z_k z_l] - \sum_{k=1}^{M}\sum_{l=1}^{M}w_{jk} w_{jl} \mathbb{E}[z_k] \mathbb{E}[z_l].
\end{align}
Furthermore, if we divide the first term into diagonalized and non-diagonalized terms,
\begin{align}
\label{eq:var4}
\mathbb{V}\biggl[\sum_{k=1}^{M}w_{jk} z_k\biggr] & = \sum_{k=1}^{M}w_{jk}^2\mathbb{E}[z_k^2] + \sum_{\substack{k,l=1 \\ k\ne l}}^{M}w_{jk}w_{jl}\mathbb{E}[z_k z_l] - \sum_{k=1}^{M}\sum_{l=1}^{M}w_{jk} w_{jl} \mathbb{E}[z_k] \mathbb{E}[z_l].
\end{align}

Let us assume that $\boldsymbol{z}$ has been normalized \update{by BN} to have a mean of 0 and a variance of 1, which implies that $\mathbb{E}[z] = 0$ and $\mathbb{E}[z^2] = 1$. Therefore, Eq. \eqref{eq:var4} can be transformed as follows: 
\begin{align}
\label{eq:var5}
    \mathbb{V}\biggl[\sum_{k=1}^{M}w_{jk} z_k\biggr] & = \sum_{k=1}^{M}w_{jk}^2 + \sum_{k \neq l}w_{jk}w_{jl}\mathbb{E}[z_k z_l].
\end{align}
Comparing Eqs. \eqref{eq:var4} and \eqref{eq:var5}, it can be observed that the influence of $z$ disappears from the first term of Eq. \eqref{eq:var4}, and the third term becomes zero by introducing the normalization. As a result, it can be stated that the impact of $z$ on the variance has been significantly reduced.

Eq. \eqref{eq:var5} indicates that the variance is affected by the covariance of $\boldsymbol{z}$ without being influenced by $\boldsymbol{z}$ itself. However, the effect of the non-diagonal elements is generally not significant because correlations between feature maps are low when $M$ is sufficiently large. Based on the above, we hypothesize that we can mitigate the effects of $\hat{y}_j$ and $\boldsymbol{z}$ by applying normalization to $\boldsymbol{z}$. 

\subsection{\update{Task correction by $\gamma \log{(\text{csg})} + \delta \log{(\text{cn})}$}}
\label{sec:tasks}
In Eq. \eqref{eq:var5}, it was demonstrated that the insertion of BN can reduce the influence of $z$ on the variance. However, while this influence is expected to be relatively minor, the second term of Eq. \eqref{eq:var5} still retains some dependency on $z$. Therefore, we hypothesize that the influence appearing in the second term is affected by factors such as task difficulty and the number of output classes. To address this, we aim to reduce the impact by extending the temperature determination coefficients, as shown in Eq. \eqref{eq:final}.

\update{Note that the second term in Eq.~\eqref{eq:var5} merely indicates that some inter–feature covariance may persist after BN, but it does not guarantee a one–to–one correspondence between that covariance and task difficulty. Indeed, when the latent representations are sufficiently separated, $\mathbb{E}[z_k z_l]\!\approx\!0$ can still hold even for hard tasks. To capture both cases in a unified way, we incorporate two complementary correction viewpoints:}

\begin{enumerate}
  \item[\textbf{(a)}] \emph{Representation-space correction:}
        \update{Empirically, tasks with highly similar classes exhibit larger inter–feature correlations.  We therefore adopt the CSG as a proxy for this effect (see Appendix~\ref{app:csg_corr_en} for a detailed derivation).}
  \item[\textbf{(b)}] \emph{Softmax-gradient correction:}
        \update{As the number of classes $C$ increases, the expected true-class probability approaches $1/C$, diluting the cross-entropy gradient $\partial L / \partial z_i$. This gradient dilution can be compensated by reducing
        the temperature~$T$ (see Appendix~\ref{app:cn_corr_en} for a detailed derivation).}
\end{enumerate}

\update{Combining these two viewpoints yields the final temperature correction rule
\[
   \;\gamma\log(\text{csg})\;+\;\delta\log (\text{cn}),
\]
where $\gamma>0$ corrects for the increase in inter–feature correlation (i.e.\ task difficulty) and $\delta<0$ offsets the softmax dilution caused by
larger class counts.}


\section{Effect of BN insertion}
\label{sec:bnexp}
In the subsequent sections, we first examine the impact of BN insertion in Section \ref{sec:bnexp}, then optimize the temperature determination coefficients in Section \ref{sec:optim}, and finally verify the effectiveness of the newly derived temperature determination method in Section \ref{sec:exp}.

In these experiments, we investigate the effect of inserting BN in image classification tasks. The research questions are as follows: 

\begin{enumerate}
    \item How does the insertion of BN just before the output layer affect estimation accuracy?
    \item Does the insertion of BN just before the output layer stabilize the temperature determination coefficients?
\end{enumerate}

\subsection{Experimental settings}
We utilized the CIFAR10/100 \cite{cifar10}, STL10 \cite{stl10}, and Tiny ImageNet (Tiny IN) \cite{imagenet} datasets and employed VGG9 \cite{vgg}, ResNet10 \cite{resnet}, and PyramidNet10 \cite{pyramid} as model architectures. Due to resource constraints and the need for numerous trials, we selected relatively small datasets and shallow model architectures. 

Our experiments utilized various models to equip the classifier $f$ with one fully connected layer. For all models, we initialized the weights of the feature extractor with the initialization proposed by He \textit{et al.} \cite{init} and the weights of the classifier with a uniform distribution $U(-1/\sqrt{M},1/\sqrt{M})$. We did not use bias terms in the convolutional layer. For model training, we used stochastic gradient descent (SGD) with a momentum of 0.9 and a weight decay of 0.0005. We set the initial learning rate to 0.1 and employed cosine annealing \cite{cosine_annealing} as the learning rate scheduler. The number of epochs and batch size were set to 200 and 128, respectively. We applied standardization as preprocessing and employed cropping, horizontal flip, and rotation at random. 

We increased or reduced $M$ by uniformly adjusting the number of filters in the model. For example, standard ResNet includes [64, 128, 256, 512] filters for each block. We reduced $M$ by dividing all filters by a factor of $n$. We report the median accuracy for multiple trials using different random seeds. 

\subsection{Effects of temperature in image recognition}
\begin{figure}[h]
\centering
 \includegraphics[width=0.5\textwidth]
      {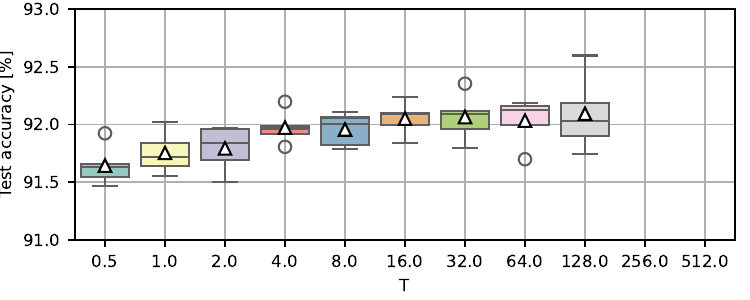}
 \caption{Change in accuracy relative to $T$ in the CIFAR10 environment using VGG9 with $M = 512$. The results for $T = 256$ and $T = 512$ are below the drawing range.}
 \label{fig:box}
\end{figure}

In Fig. \ref{fig:box}, we present the evaluation results of VGG9 ($M = 512$) on CIFAR-10. The performance improves as $T$ increases up to $T = 128$, after which it declines rapidly ($T\geq256$ is outside the drawing range). As displayed in Fig. \ref{fig:softmax}, lower values of $T$ lead to sharper $\hat{\boldsymbol{y}}$. Setting $T$ too low can cause the model to rapidly converge to local minima. Conversely, higher values of $T$ lead to flatter $\hat{\boldsymbol{y}}$, allowing the model to explore a wider parameter space. Therefore, setting $T$ to an excessively high value may prevent the model from converging. 

\begin{figure}[h]
    \centering
    \includegraphics[width=1.\textwidth]{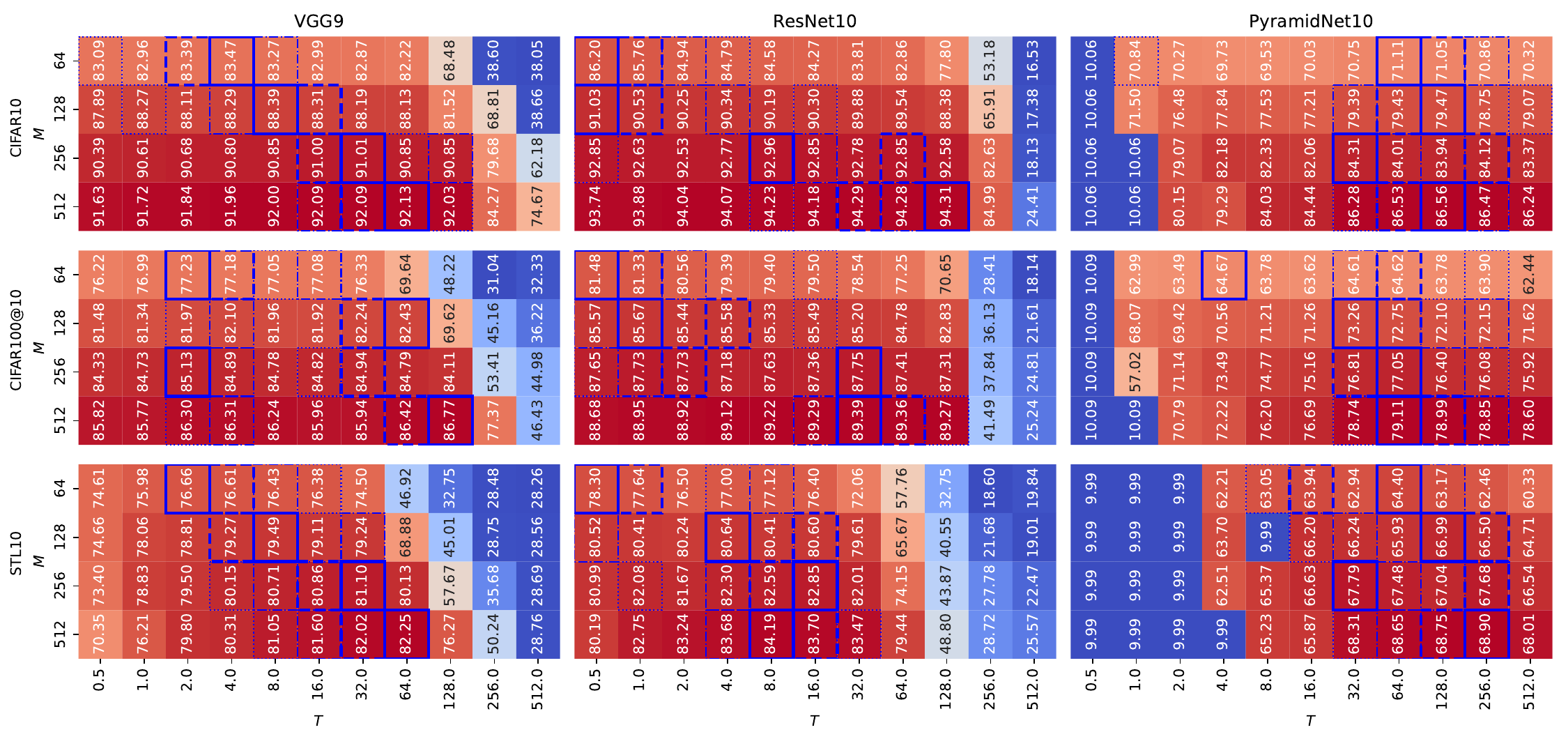}
    \caption{Test accuracies [\%] for each temperature parameter in various scenarios (without insertion of the normalization layer). Only CIFAR-100@10 has 10 classes extracted from its superclasses (using only even class numbers) to standardize the number of output units to 10.}
    \label{fig:normal}
\end{figure}

In Fig. \ref{fig:normal}, we present results across various model architectures and datasets with varying temperature parameters $T$. To standardize the CN, we extracted a subset of CIFAR100 containing superclasses at even positions (10 classes, 25,000 samples, denoted as CIFAR100@10). In each subfigure, the $x$-axis represents $T$, the $y$-axis represents $M$ for each model architecture, and the intensity represents the median test accuracy. A solid line denotes the best results, while a dashed line denotes the second-best results. From Fig. \ref{fig:normal}, we observe the trend that $\hat{T}^*$ increases as $M$ increases. However, as reported in 
the related study \cite{hase}, this trend varies across datasets and model architectures. 

\subsection{Effect of BN insertion}
\begin{figure}[h]
    \centering
    \includegraphics[width=1.\textwidth]{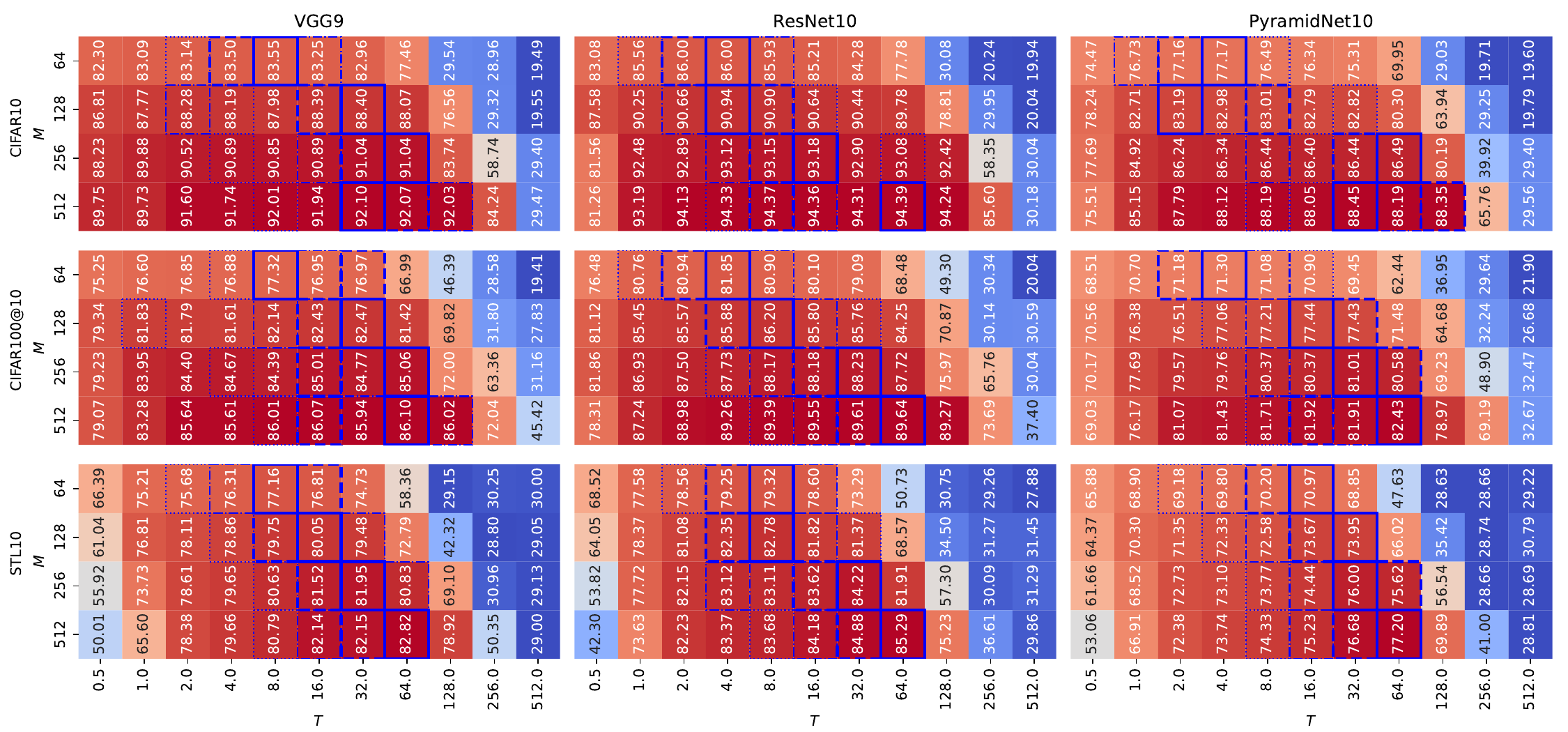}
    \caption{Test accuracies [\%] for each temperature parameter in various scenarios (with BN insertion). Only CIFAR-100@10 has 10 classes extracted from its superclasses (using only even class numbers) to standardize the number of output units to 10.}
    \label{fig:bn}
\end{figure}

In Fig. \ref{fig:bn}, we present the results of inserting BN just before the output layer. This figure can be interpreted in the same way as Fig. \ref{fig:normal}. The trend of the optimal temperature parameter appears more robust than that displayed in Fig. \ref{fig:normal}. Notably, at the point where the performance declines in the region of high values of $T$, BN insertion causes the trend to be similar across different model architectures. 

\begin{figure}[h]
    \centering
    \includegraphics[width=1.\textwidth]{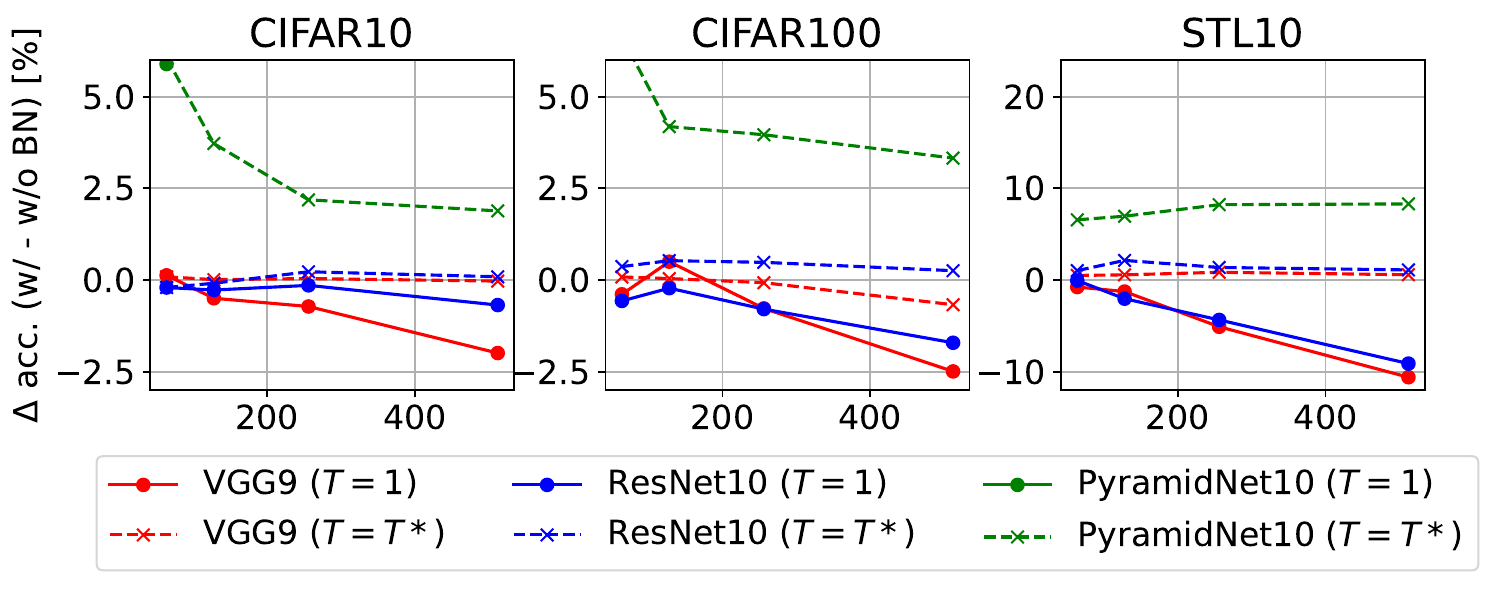}
    \caption{Accuracy difference [\%] between scenarios with and without BN insertion. Each value is calculated by subtracting the accuracy without BN from the accuracy with BN.}
    \label{fig:diff}
\end{figure}

Next, we analyze the effect of BN insertion on estimation accuracy. In Fig. \ref{fig:diff}, we illustrate the differences in estimation accuracy with and without BN insertion (as $\Delta$ acc.). Positive values indicate that BN insertion improves performance, while negative values indicate that it reduces performance. As shown by the solid line, inserting BN at $T = 1$ shows that the accuracy decreases with increasing M. However, by combining $T = \hat{T}^*$ and BN insertion, the accuracy improves in many cases. The performance improvement is particularly notable for a relatively high-$M$ and difficult dataset (STL-10). 

Based on the experimental results, inserting BN just before the output layer can provide a more robust estimation of the temperature determination coefficients across different datasets and model architectures. We infer the reason many conventional methods do not insert BN just before the output layer is that, with the default setting of $T = 1$, accuracy decreases due to BN insertion. However, it is possible to achieve a more accurate prediction once the temperature parameter $T$ is set appropriately. These results highlight the importance of optimizing the temperature parameter $T$.

\section{Optimization of Temperature Determination Coefficients}
\label{sec:optim}
In this section, we estimate the temperature determination coefficients through optimization. We experimentally demonstrate that (i) we can stabilize the estimation of the temperature determination coefficients by BN insertion and (ii) we can improve the estimation accuracy by optimizing the temperature parameter $T$. Based on these two results, we further introduce (iii) a correction of temperature using CSG and CN. 

\subsection{Estimation of performance for each $T$ and optimization of $T$}
\begin{figure}[h]
    \centering
    \includegraphics[width=\textwidth]{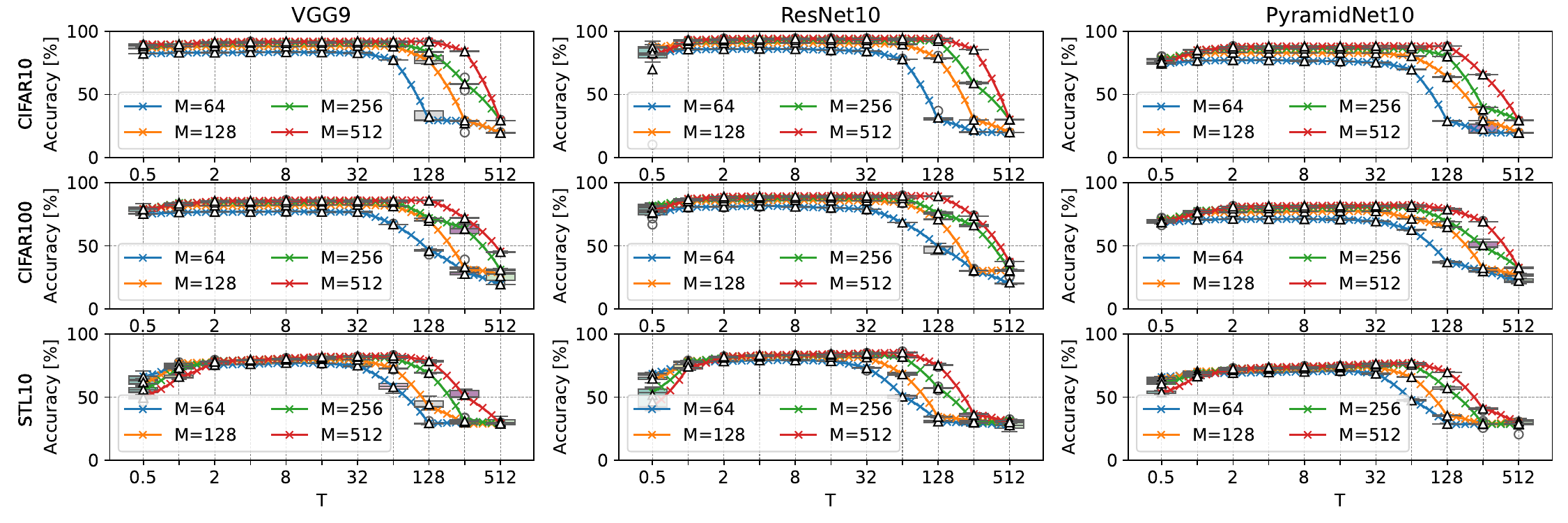}
    \caption{Estimation of test accuracy [\%] for each temperature parameter (with BN insertion). The boxplots represent the observed values, and triangles indicate the median values. The solid line plots, marked with $\times$, represent the estimated values obtained through linear interpolation of the observed mean values.}
    \label{fig:interp}
\end{figure}

Since the optimal temperature $T^*$ under given conditions is unknown, we estimate the temperature determination coefficients $\alpha$ and $\beta$ as follows:
\begin{itemize}
    \item[(i)] We estimate the test performance at unknown $T$ by linear interpolation (Fig. \ref{fig:interp}). 
    \item[(ii)] We determine the optimal temperature using $\hat{T}^{*}=\alpha \sqrt{M} + \beta$. 
    \item[(iii)] By maximizing $\sum_{c \in C}\operatorname{interpolation}_c(\hat{T}^*)$, we estimate the temperature determination coefficients $\alpha$ and $\beta$. 
\end{itemize}

In step (i), as illustrated in Fig. \ref{fig:interp}, we estimate the test performance at unknown $T$ by linear interpolation based on the results in Fig. \ref{fig:bn}. We present the results of Fig. \ref{fig:bn} using a boxplot and draw a solid line indicating the test performance at unknown $T$ that is linearly interpolated. As the test performance does not significantly change with slight changes in $T$, we consider linear interpolation to be sufficient. This allows us to estimate the test performance in the range $0 < T \leq 512$. 
In step (ii), as mentioned in Section \ref{sec:hypothesis}, we determine the optimal temperature $T^*$ by $\hat{T}^{*} = \alpha \sqrt{M} + \beta$. In step (iii), we estimate the temperature determination coefficients by maximizing the sum of the estimated test accuracy under various conditions $C$. We denote the estimated test accuracy for a given condition $C$ and temperature parameter $T$ as $\operatorname{interpolation}_c(T)$. For example, we can estimate the optimal test accuracy of VGG9 on CIFAR-10 by setting the condition $C$ as $M = [64, 128, 256, 512]$. We employ differential evolution for optimization \cite{de}.

Table \ref{table:estimated} displays the temperature determination coefficients estimated in each environment in Fig. \ref{fig:interp} along with the test accuracy using these coefficients. We observe that, on average, $\alpha = 4$ and $\beta = -25$ with and without BN. However, the standard deviation of the estimated coefficient $\alpha$ is 2.17 without BN, which decreases to 1.86 with BN. Therefore, the stability of the coefficients improves with BN insertion. From the perspective of test performance, the test accuracy ($\mu_{acc}$) of the optimal temperature $\hat{T}^*$ with BN is superior to that without BN. These results validate the effectiveness of our proposed method. 
\begin{table}[h] 
\centering
    \caption{Temperature determination coefficient estimated through optimization and the corresponding average accuracy [\%].}
    \label{table:estimated}
    \begin{tabular}{l|ccc|ccc|c}
    \toprule
 &	\multicolumn{3}{c|}{w/o BN} &	\multicolumn{3}{c|}{w/ BN} &	 \\ 
 &	$\alpha$ &	$\beta$ &	$\mu_{acc}$ &	$\alpha$ &	$\beta$ &	$\mu_{acc}$ &	$\Delta_{acc}$ \\ \midrule
\multicolumn{8}{l}{\textbf{CIFAR10}} \\
\quad VGG9 &	2.09  &	-12.70  &	88.73  &	3.42  &	-19.32  &	88.76  &	0.03  \\ 
\quad ResNet10 &	7.62  &	-60.92  &	90.74  &	0.84  &	-4.29  &	91.11  &	0.37  \\ 
\quad PyramidNet10 &	1.98  &	-13.84  &	83.69  &	7.09  &	-52.70  &	83.69  &	0.01  \\  \midrule
\multicolumn{8}{l}{\textbf{CIFAR100}} \\
\quad VGG9 &	7.63  &	-58.14  &	82.71  &	6.69  &	-45.52  &	82.71  &	0.00  \\ 
\quad ResNet10 &	3.79  &	-30.07  &	85.96  &	2.68  &	-17.42  &	86.41  &	0.45  \\ 
\quad PyramidNet10 &	3.41  &	-13.13  &	77.93  &	4.11  &	-29.07  &	78.03  &	0.09  \\  \midrule
\multicolumn{8}{l}{\textbf{STL10}} \\
\quad VGG9 &	3.10  &	-22.80  &	79.78  &	2.95  &	-15.30  &	80.41  &	0.63  \\ 
\quad ResNet10 &	1.41  &	-11.17  &	81.14  &	3.18  &	-21.39  &	82.62  &	1.48  \\ 
\quad PyramidNet10 &	4.64  &	-20.47  &	74.08  &	2.98  &	-7.87  &	74.46  &	0.38  \\ \midrule
\quad Avg. &	3.96  &	-27.03  &	82.75  &	3.77  &	-23.65  &	83.13  &	0.38  \\ 
\quad S.D. &	2.17  &	18.29  &	4.95  &	\textbf{1.86}  &	\textbf{15.30}  &	4.89  &	0.44  \\ 
    \bottomrule
    \end{tabular}
\end{table}

\subsection{Effect of the task}
Revisiting Fig. \ref{fig:bn}, we can observe that while BN insertion can suppress the effect of differences in model architecture, the effect of differences in dataset slightly remains. We hypothesized that task difficulty causes this phenomenon. As the number of similar classes increases (i.e., the task difficulty increases), the greater the impact on learning since the output distribution is gradually flattened with increasing $T$, as illustrated in Fig. \ref{fig:softmax}. Therefore, we analyzed changes in the temperature determination coefficients with respect to task difficulty. We constructed datasets with different task difficulties by extracting subsets containing 10 classes from CIFAR100 (referred to as the CIFAR100 subset). From these subsets, we utilized two relatively difficult subsets and two relatively easy subsets. Moreover, we simultaneously utilized the original CIFAR-100 and Tiny IN datasets. To calculate task difficulty, we employed the CSG \cite{Frederic2019}. We computed the CSG scale using the original images. While the original implementation trains an autoencoder and employs t-SNE, we do not use either of them, but it is sufficient for this case. For CSG calculation, we employed the publicly available implementation, which can be found at \url{https://github.com/Dref360/spectral-metric}. 

\begin{figure}[h]
\centering
 \includegraphics[width=\textwidth]
      {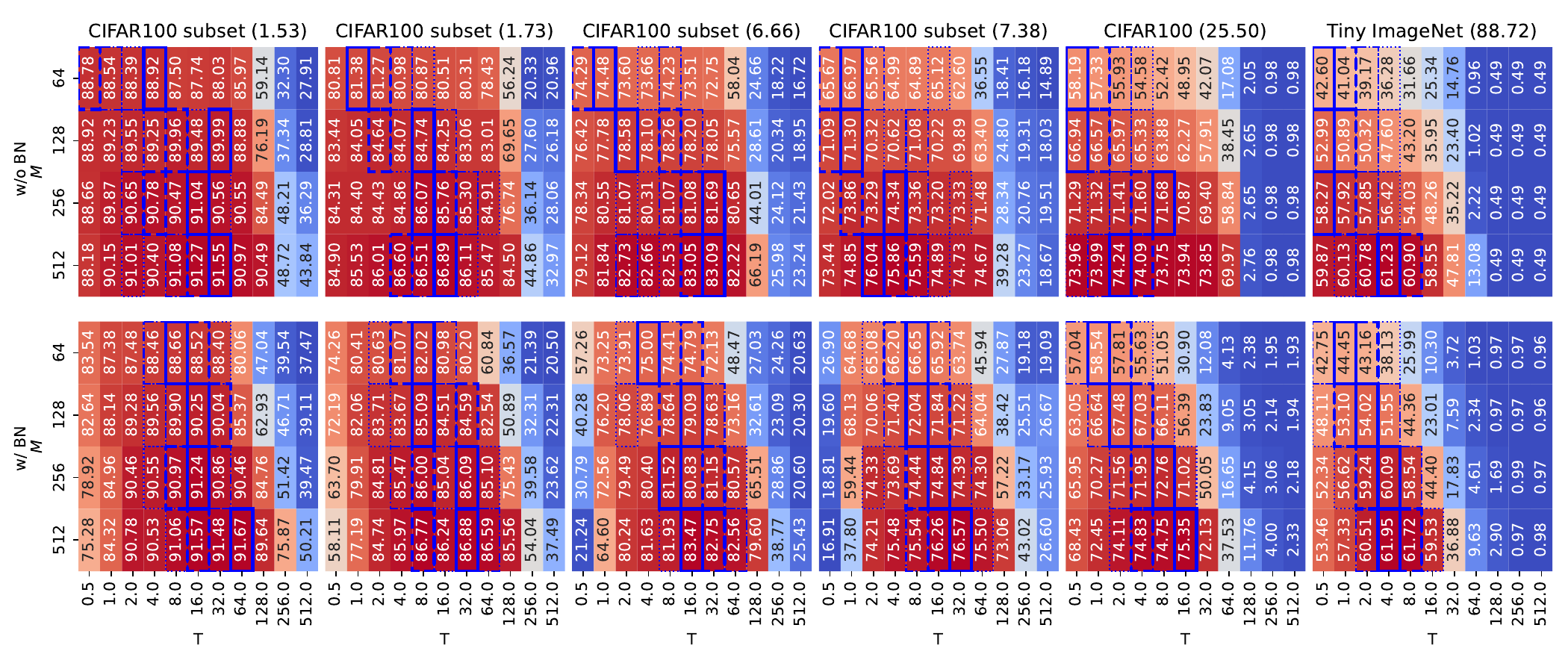}
 \caption{Test accuracies [\%] for each temperature parameter in different task difficulty scenarios using ResNet10. The upper row displays results without BN insertion (conventional), while the lower row displays results with BN insertion (proposed method). Values in brackets indicate CSGs.}
 \label{fig:csg}
\end{figure}

In Fig. \ref{fig:csg}, we present the results on four subsets of CIFAR100, the original CIFAR100, and Tiny IN, employing ResNet10 as the model architecture. We observe the following trends, which are consistent with the previous section:
\begin{enumerate}
    \item BN insertion stabilizes the temperature determination coefficients.
    \item While BN insertion reduces the test accuracy at $T = 1$, it improves the test accuracy at $T = \hat{T}^*$.
    \item Higher task difficulty scores (CSG) correlate with greater improvement due to temperature adjusting. 
\end{enumerate}
Furthermore, we observe that as the CSG increases, the optimal temperature $T^*$ decreases. An increase in the CN may be responsible for the decrease in $T^*$. These findings are consistent with the hypothesis that increased $T$ has a larger impact on similar classes, particularly in challenging tasks. 

Based on the above, the optimal temperature determination coefficients can be estimated by inserting BN, representing $\hat{T}^*$ as a function of $M$, and correcting with CSG and CN. Based on Eq. \eqref{eq:final}, we defined as follows:

\begin{equation}
    \hat{T}^* = \operatorname{clip}(\alpha \sqrt{M} + \beta + \gamma \log(csg) + \delta \log(cn), \epsilon, 512),
\end{equation}

In the case without correction, $\gamma = \delta = 0$. When only the CSG is used for correction (CSG Corr.), $\delta = 0$, while when only the CN is used for correction (CN Corr.), $\gamma = 0$. The ${\rm clip}()$ function represents a clipping operation that limits the values between $\epsilon$ and $512$. We experimentally set the upper bound of $T$ to $512$ and the lower bound to $1$. 

As in the previous section, we performed optimization by cross-validation (CO) to estimate the temperature determination coefficients. Table \ref{table:csgcorr} displays the test accuracy obtained by performing linear interpolation on each dataset. We observe a small improvement in overall performance, particularly notable on the Tiny IN dataset. For the original CIFAR100 dataset, while the performance decreases slightly, it can be improved by increasing the number of samples for simulation and optimization. 
\begin{table}[h]
\centering
    \caption{Average estimation accuracy [\%] achieved by our proposed methods. Each value indicates the average test accuracy for all $M$.}
    \label{table:csgcorr}
    \begin{tabular}{@{}l|cccc|cc|c@{}}
    \toprule
\multicolumn{1}{r|}{Dataset} &	\multicolumn{4}{c|}{CIFAR100 subset} &	CIFAR100 &	Tiny IN &	\multirow{3}{*}{Avg.} \\
\multicolumn{1}{r|}{CSG} &	1.53 &	1.73 &	6.66 &	7.38 &	25.50 &	88.72 &	\\ 
\multicolumn{1}{r|}{CN} &	10 &	10 &	10 &	10 &	100 &	200 &	\\\midrule
ResNet 9 w/ BN (CO) &	89.75  &	84.12  &	78.41  &	71.15  &	68.16  &	49.55  &	73.52  \\ \midrule
 \quad+ CSG Corr. (CO) &	90.26  &	84.66  &	79.18  &	72.05  &	66.43  &	54.48  &	74.51  \\ 
 \quad+ CN Corr. (CO) &	90.31  &	84.51  &	79.29  &	72.29  &	67.56  &	54.30  &	74.71  \\ 
 \quad+ CSG\&CN Corr. (CO) &	90.20  &	84.61  &	79.48  &	72.33  &	66.58  &	54.66  &	74.64  \\ \midrule
 \quad+ CSG Corr. (GO) &	90.32  &	84.65  &	79.32  &	72.20  &	67.58  &	54.90  &	74.83 \\ 
 \quad+ CN Corr. (GO) &	90.24  &	84.76  &	79.38  &	72.29  &	67.87  &	55.01  &	74.92  \\ 
 \quad+ CSG\&CN Corr. (GO) &	90.36  &	84.74  &	79.39  &	72.23  &	68.14  &	55.03  &	74.98  \\ 
        \bottomrule
    \end{tabular}
\end{table} 

Table \ref{table:csgcorr} also displays the results of performing global optimization (GO) on all experimental results displayed in Fig. \ref{fig:csg}. As expected, the average performance exceeds that obtained by optimization using cross-validation. While performing correction is beneficial, the difference in performance between CSG Corr. and CN Corr. is negligible. Therefore, one can simply use CN Corr. or CSG \& CN Corr. when the CSG value can be computed. Finally, we can estimate the optimal temperature parameter $\hat{T}^*$ based on the temperature determination coefficients estimated by global optimization, as demonstrated in Table \ref{table:coeff}. 

\begin{table}[h]
\centering
    \caption{Temperature determination coefficients obtained by global optimization.}
    \label{table:coeff}
    \begin{tabular}{@{}l|cccc@{}}
    \toprule
 &	$\alpha$ &	$\beta$ &	$\gamma$ &	$\delta$ \\ \midrule
$\hat{T}^*$ &	0.7239 &	-4.706 &	 &	 \\ 
$\hat{T}^*_{csg}$ &	0.4111 &	6.848 &	-2.024 &	 \\ 
$\hat{T}^*_{cn}$ &	0.4051 &	6.656 &	 &	-1.973 \\ 
$\hat{T}^*_{csgcn}$ &	0.3192 &	20.74 &	3.746 &	-7.38 \\ 
        \bottomrule
    \end{tabular}
\end{table}

\section{Effectiveness validation with real problems}
\label{sec:exp}
In the previous section, we discussed the performance of temperature parameters based on estimated test accuracy through linear interpolation. We also employed relatively shallow model architectures. To validate the robustness of our proposed method, we estimate the optimal temperature $\hat{T}^*$ for standard deep learning models. Based on the empirical data obtained during actual training and validation, we evaluate the effectiveness of our proposed method. The training settings are identical to those described in the previous section, with the exception of batch size, which varies across datasets. 

We compared the following approaches:
\begin{enumerate}
    \item $\boldsymbol{T = 1}$: The conventional method that uses $T = 1$ without BN insertion (current \textbf{default} setting). 
    \item $\boldsymbol{T = \sqrt{M}}$: The method used in self-attention \cite{Vaswani2017}, which employs $T=\sqrt{M}$ without BN insertion. 
    \item \textbf{Insert LN}: The method that inserts LN just before applying the softmax function without BN insertion \cite{hase}.
    \item \textbf{Ours ($\hat{T}^*$)}: The method that inserts BN just before the output layer and employs $\hat{T}^*$. 
    \item \textbf{Ours ($\hat{T}^*_{csg}$)}: The method that corrects the temperature parameter to $\hat{T}^*_{csg}$ calculated from the CSG with global optimization. 
    \item \textbf{Ours ($\hat{T}^*_{cn}$)}: The method that correct the temperature parameter to $\hat{T}^*_{cn}$ based on the CN with global optimization. 
    \item \textbf{Ours ($\hat{T}^*_{csgcn}$)}: The method that corrects the temperature parameter to $\hat{T}^*_{csgcn}$ using the CSG and CN with global optimization. 
\end{enumerate}

\subsection{Evaluation of robustness to differences in model architecture}

We validated our method on various model architectures, including ConvNets and ViTs. 

\subsubsection{Convolutional neural networks}
We employed the following widely used ConvNets architectures: ResNet50 \cite{resnet}, EfficientNet (Eff. Net) b0/b5 \cite{pmlr-v97-tan19a}, RegNet8GF \cite{Radosavovic_2020_CVPR}, and ConvNeXt-Small \cite{Liu_2022_CVPR}. We trained all models from scratch on CIFAR10 and CIFAR100. 

In Table \ref{table:actualmodels}, we present the performance of each model. Notably, our proposed method outperforms other methods in many cases. Interestingly, $T = \sqrt{M}$ derived from $1/\sqrt{d_k}$ from the self-attention study \cite{Vaswani2017} sometimes outperforms our method. This method differs from our proposed method in terms of BN insertion and the calculation of $T$. Since $T = \sqrt{M}$ does not significantly differ from our proposed method in some cases, it is expected to exhibit high performance, particularly for datasets with a low CSG. 
However, it is sensitive to changes in dataset and model architecture, resulting in significantly lower performance in cases such as ConvNeXt on CIFAR10 and Eff. Net b5 on CIFAR100. In terms of average performance, our method achieves the best performance on CIFAR10, while the addition of CSG Corr. achieves the best performance on CIFAR100. It should be noted that because CSG correction aims to improve robustness against difficult datasets such as Tiny ImageNet, it may not be as effective for easier datasets such as CIFAR10/100. 
\begin{table}[h] 
\centering
    \caption{Performance evaluation for each Torchvision model in actual measurements [\%].}
    \label{table:actualmodels}
    \begin{tabular}{@{}l|ccccc|c@{}}
    \toprule
  &	ResNet50 &	Eff. Net b0 &	Eff. Net b5 &	RegNet 8 GF &	ConvNeXt-S &	Avg. \\ \midrule
\textbf{CIFAR10} &  &  &  &  &  \\
\quad$T = 1$ &	84.80 &	87.10 &	88.83 &	88.72 &	70.02 &	83.89 \\ 
\quad$T = \sqrt{M}$ &	\ub{88.70} &	87.51 &	\ub{91.01} &	\ub{91.18} &	68.07 &	85.29 \\ 
\quad Insert LN \cite{hase} &	85.66 &	87.45 &	81.27 &	88.97 &	70.03 &	82.68 \\ 
\quad Ours ($\hat{T}^{*}$) &	88.57 &	\ux{87.61} &	\ux{90.79} &	\ux{90.74} &	\ub{77.94} &	\ub{87.13} \\ 
\quad Ours ($\hat{T}^{*}_{csg}$) &	\ux{88.62} &	87.41 &	90.61 &	90.47 &	\ux{76.68} &	\ux{86.76} \\ 
\quad Ours ($\hat{T}^{*}_{cn}$) &	88.45 &	87.60 &	90.15 &	90.23 &	71.87 &	85.66 \\ 
\quad Ours ($\hat{T}^{*}_{csgcn}$) &	88.46 &	\ub{87.72} &	89.75 &	90.30 &	71.07 &	85.46 \\ \midrule
\textbf{CIFAR100} &  &  &  &  &  \\
\quad$T = 1$ &	49.18 &	58.75 &	62.34 &	56.52 &	\ub{50.34} &	55.43 \\ 
\quad$T = \sqrt{M}$ &	\ub{60.99} &	46.23 &	15.97 &	\ub{65.16} &	32.62 &	44.19 \\ 
\quad Insert LN \cite{hase} &	49.64 &	58.48 &	55.50 &	55.41 &	\ux{49.58} &	53.72 \\ 
\quad Ours ($\hat{T}^{*}$) &	\ux{59.95} &	59.01 &	66.05 &	\ux{64.94} &	35.25 &	57.04 \\ 
\quad Ours ($\hat{T}^{*}_{csg}$) &	58.70 &	\ux{59.20} &	\ub{68.20} &	63.60 &	41.70 &	\ub{58.28} \\ 
\quad Ours ($\hat{T}^{*}_{cn}$) &	57.93 &	58.94 &	\ux{67.81} &	63.48 &	41.00 &	57.83 \\ 
\quad Ours ($\hat{T}^{*}_{csgcn}$) &	57.89 &	\ub{60.14} &	65.93 &	63.56 &	42.70 &	\ux{58.04} \\ 
        \bottomrule
    \end{tabular}
\end{table} 

\subsubsection{Vision transformers}
We also evaluated our methods on recently developed ViT-based architectures: ViT-B/16 \cite{dosovitskiy2020vit} and Swin Transformer Tiny (SwinT) \cite{swint}. We utilized Caltech101 \cite{caltech} as the dataset. 

In Table \ref{table:actualtransformer}, we present the results for both models on the Caltech101 dataset. For both settings, our proposed method achieves the best performance. Although the performance significantly decreases with CSG and CN correction, it may be improved through individual hyperparameter tuning. For comparative experiments in a unified environment, we present the results as obtained.

\subsection{Evaluation of robustness to dataset differences}
We evaluated whether our proposed method was effective for various image classification datasets that were not used for optimization. We used the following popular vision datasets including fine-grained image recognition: German Traffic Sign Recognition Benchmark (GTS)\cite{gtsrb}, EuroSAT (SAT)\cite{helber2019eurosat}, ISIC Challenge 2019 (ISIC)\cite{Tschandl2018, Codella2018, Hernandez2024}, Caltech101 (CT)\cite{caltech}, Flowers102 (FLW)\cite{flowers}, Oxford-IIIT Pet (PET)\cite{pet}, Describable Textures Dataset (DTD)\cite{dtd}, CUB-2011-200 dataset (CUB)\cite{WahCUB_200_2011}. The validation was conducted under the following three scenarios: (1) training ResNet50 from scratch (init), (2) fine-tuning a ResNet50 pre-trained on ImageNet-1k (FT), and (3) fine-tuning a SwinT model pre-trained on ImageNet-1k (FT). During the fine-tuning phase, both models were trained for 50 epochs using a maximum learning rate of 0.01.
\begin{table}[h] 
\centering
    \caption{Performance evaluation of transformer models on the Caltech101 dataset in actual measurements [\%].}
    \label{table:actualtransformer}
    \begin{tabular}{@{}l|cc}
    \toprule
 &	ViT B/16 &	SwinT \\ \midrule
$T = 1$ &	50.38 &	54.07 \\ 
$T = \sqrt{M}$ &	47.54 &	61.29 \\ 
Insert LN \cite{hase} &	49.92 &	68.20 \\ 
Ours ($\hat{T}^{*}$) &	54.15 &	70.66 \\ 
Ours ($\hat{T}^{*}_{csg}$) &	\ux{54.30} &	\ux{72.77} \\ 
Ours ($\hat{T}^{*}_{cn}$) &	\ub{55.34} &	\ub{74.46} \\ 
Ours ($\hat{T}^{*}_{csgcn}$) &	5.03 \tablefootnote{53.61\% when the maximum learning rate is changed to 0.01} &	40.51 \tablefootnote{73.52\% when the maximum learning rate is changed to 0.01} \\ 
        \bottomrule
    \end{tabular}
\end{table} 

In Table \ref{table:actualdata}, we present the results for each dataset. Despite each correction formula $\hat{T}^*$ to $\hat{T}^*_{csgcn}$ being derived from CIFAR10/100 and Tiny ImageNet, our proposed method exhibits superior performance. Specifically, corrections based on CN or CSG\&CN are effective in numerous instances, illustrating the broad applicability of our method to common vision tasks---even when those tasks were not used in the optimization of temperature determination coefficients. 
\begin{table}[h] 
\centering
    \caption{Performance evaluation of three scenarios for each dataset in actual measurements [\%].}
    \label{table:actualdata}
    \begin{tabular}{@{}l|cccccccc|c@{}}
    \toprule
\multicolumn{1}{r|}{Dataset} &	GTS &	SAT &	ISIC &	CT &	FLW &	PET &	DTD &	CUB &	\multirow{3}{*}{Avg.} \\ 
\multicolumn{1}{r|}{CSG} &	1.39 &	3.08 &	3.85 &	5.59 &	7.47 &	22.80 &	25.86 &	75.32 &	 \\ 
\multicolumn{1}{r|}{CN} &	43 &	10 &	8 &	101 &	102 &	37 &	47 &	200 &	 \\ \midrule
\textbf{ResNet50 init} &	 &	 &	 &	 &	 &	 &	 &	 &	 \\ 
\quad $T=1$ &	70.16 &	97.33 &	\ub{76.16} &	83.49 &	39.93 &	55.74 &	34.10 &	51.98 &	63.61 \\ 
\quad $T=\sqrt{M}$ &	88.69 &	97.56 &	74.07 &	86.37 &	39.47 &	76.02 &	34.63 &	35.42 &	66.53 \\ 
\quad Insert LN \cite{hase} &	71.56 &	96.85 &	\ux{75.17} &	87.63 &	48.84 &	71.46 &	36.65 &	54.92 &	67.89 \\ 
\quad Ours ($\hat{T}^{*}$) &	87.26 &	97.59 &	66.57 &	87.52 &	53.54 &	\ux{76.12} &	\ub{42.29} &	63.01 &	71.74 \\ 
\quad Ours ($\hat{T}^{*}_{csg}$) &	87.56 &	\ux{97.69} &	72.16 &	88.63 &	\ub{54.81} &	\ub{77.49} &	\ux{36.76} &	60.08 &	71.90 \\ 
\quad Ours ($\hat{T}^{*}_{cn}$) &	\ux{92.01} &	\ub{97.90} &	70.03 &	\ub{90.59} &	\ux{54.76} &	74.03 &	35.53 &	\ub{64.29} &	\ub{72.39} \\ 
\quad Ours ($\hat{T}^{*}_{csgcn}$) &	\ub{92.18} &	97.40 &	69.32 &	\ux{90.05} &	52.98 &	75.63 &	36.28 &	\ux{63.34} &	\ux{72.15} \\ \midrule
\textbf{ResNet50 FT} &	 &	 &	 &	 &	 &	 &	 &	 &	 \\ 
\quad $T=1$ &	89.14 &	98.44 &	\ux{84.72} &	\ub{96.43} &	90.18 &	87.44 &	62.98 &	75.03 &	85.54 \\ 
\quad $T=\sqrt{M}$ &	87.96 &	\ux{98.52} &	82.75 &	33.37 &	17.30 &	77.19 &	34.79 &	5.14 &	54.63 \\ 
\quad Insert LN \cite{hase} &	91.84 &	\ub{98.58} &	\ub{84.93} &	95.31 &	\ux{90.32} &	88.09 &	63.56 &	75.13 &	85.97 \\ 
\quad Ours ($\hat{T}^{*}$) &	\ux{92.43} &	98.46 &	84.39 &	93.93 &	79.10 &	\ux{89.59} &	64.20 &	66.55 &	83.58 \\ 
\quad Ours ($\hat{T}^{*}_{csg}$) &	92.39 &	98.51 &	80.58 &	95.62 &	84.45 &	88.83 &	65.05 &	\ub{76.92} &	85.29 \\ 
\quad Ours ($\hat{T}^{*}_{cn}$) &	92.29 &	98.43 &	83.88 &	96.08 &	89.17 &	89.26 &	\ub{65.64} &	75.46 &	\ux{86.28} \\ 
\quad Ours ($\hat{T}^{*}_{csgcn}$) &	\ub{92.94} &	98.47 &	84.38 &	\ux{96.20} &	\ub{91.27} &	\ub{89.81} &	\ux{65.21} &	\ux{76.49} &	\ub{86.85} \\ \midrule
\textbf{SwinT FT} &	 &	 &	 &	 &	 &	 &	 &	 &	 \\ 
\quad $T=1$ &	89.64 &	\ub{98.57} &	\ux{84.36} &	95.85 &	\ux{80.29} &	89.40 &	65.11 &	76.77 &	85.00 \\ 
\quad $T=\sqrt{M}$ &	\ux{92.78} &	98.41 &	83.46 &	95.01 &	51.02 &	91.39 &	65.37 &	67.03 &	80.56 \\ 
\quad Insert LN \cite{hase} &	91.03 &	98.25 &	\ub{84.51} &	\ux{96.85} &	77.18 &	91.20 &	\ub{68.51} &	74.53 &	85.26 \\ 
\quad Ours ($\hat{T}^{*}$) &	92.49 &	98.41 &	80.88 &	93.74 &	66.47 &	\ux{91.71} &	66.54 &	57.40 &	80.96 \\ 
\quad Ours ($\hat{T}^{*}_{csg}$) &	92.60 &	\ux{98.43} &	83.64 &	93.93 &	60.81 &	91.14 &	\ux{68.14} &	77.99 &	83.34 \\ 
\quad Ours ($\hat{T}^{*}_{cn}$) &	\ub{93.24} &	\ux{98.43} &	80.83 &	96.66 &	76.57 &	\ub{92.15} &	67.61 &	\ux{79.19} &	\ux{85.58} \\ 
\quad Ours ($\hat{T}^{*}_{csgcn}$) &	92.13 &	\ux{98.43} &	81.32 &	\ub{96.93} &	\ub{90.21} &	91.63 &	\ux{68.14} &	\ub{79.67} &	\ub{87.31} \\ 
 \bottomrule
    \end{tabular}
\end{table} 

Across the three scenarios, no significant differences in the observed trends were detected, indicating that the proposed method remains effective regardless of changes in model architecture or the approach used for training (i.e., from initial parameters or through fine-tuning). We initially hypothesized that the proposed method might be less effective if the model had already converged near a local optimum. However, even when applying the temperature determination coefficients optimized under scratch scenarios, the results confirmed that the proposed method remained effective for transfer-learning scenarios. These findings indicate that our proposed method can be effectively combined with transfer learning, preserving its benefits even in fine-tuning scenarios.

\begin{figure}[h]
\centering
 \includegraphics[width=0.5\textwidth]
      {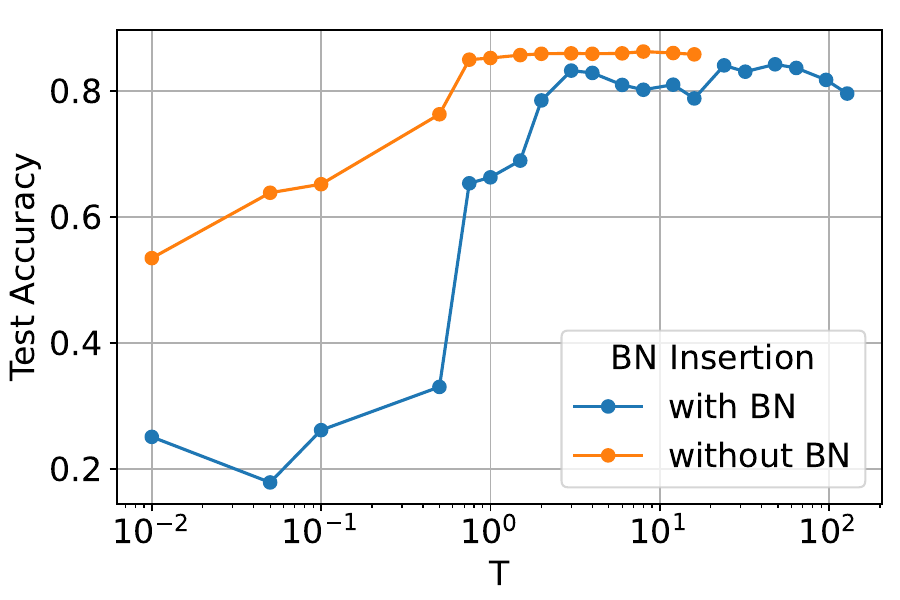}
 \caption{Test accuracies [\%] for each temperature parameter in ISIC using ResNet50.}
 \label{fig:isic}
\end{figure}

Examining the results across all datasets indicates that the proposed method generally achieves superior performance. However, it appears to be less effective, specifically on the ISIC dataset. To explore possible reasons for this, we conducted performance evaluations on the ISIC dataset by varying the temperature and comparing conditions with and without BN in the ResNet50 transfer scenario, as shown in Fig. \ref{fig:isic}. Under the with-BN condition, the proposed method achieves results close to the optimal value, $\hat{T}_{csgcn}=24.88$, confirming its efficacy. In contrast, under the without-BN condition, a wider temperature range yields high performance, resulting in comparatively poorer outcomes for the proposed method. As described in Section \ref{sec:bnexp}, incorporating BN generally improves performance when the optimal temperature is selected, although this pattern was not observed with the ISIC dataset. Further investigation is needed to clarify the conditions under which this discrepancy arises.

\subsection{Performance evaluation with the incorporation of label smoothing}
In Section \ref{sec:smooth}, we discuss the difference between label smoothing and our proposed method. However, the effect of combining label smoothing remains unclear. Therefore, we evaluated the scratch and transfer performance of ResNet50 using label smoothing.

We present the results using label smoothing in Table \ref{table:actualsmoothing}. It can be seen that our proposed method performs effectively without label smoothing. However, when incorporating label smoothing, the performance of our proposed method decreases. In comparison with Table \ref{table:actualdata}, the conventional method ($T = 1$) exhibits improved performance by incorporating label smoothing.
From the discussion in Section \ref{sec:smooth}, we expect both label smoothing and temperature scaling to play distinct roles in regularization. However, the experimental results indicate that this study's estimation formula for $T^*$ is incompatible with label smoothing. Nevertheless, since the proposed method outperforms label smoothing when $T = 1$ in terms of average performance, we conclude that it is preferable to use the proposed method rather than label smoothing.

\begin{table}[h] 
\centering
    \caption{Performance evaluation using the proposed method with label smoothing in ResNet50 [\%].}
    \label{table:actualsmoothing}
    \begin{tabular}{@{}l|ccccc|c}
    \toprule 
 &	GTS &	CT &	FLW &	PET &	DTD &	Avg. \\ \midrule
\textbf{Scratch} \\
\quad $T = 1$ &	90.20 &	89.06 &	45.03 &	59.47 &	30.16 &	62.78 \\
\quad $T = \sqrt{M}$ &	91.02 &	87.02 &	39.71 &	\ub{77.43} &	\ub{40.74} &	67.19 \\
\quad Insert LN \cite{hase} &	91.64 &	88.75 &	50.59 &	72.64 &	\ux{39.57} &	68.64 \\
\quad Ours ($\hat{T}^{*}$) &	\ux{91.97} &	\ux{90.02} &	\ub{57.42} &	\ux{76.42} &	36.70 &	\ub{70.51} \\
\quad Ours ($\hat{T}^{*}_{csg}$) &	91.27 &	\ux{90.02} &	\ux{55.68} &	75.28 &	37.82 &	70.01 \\
\quad Ours ($\hat{T}^{*}_{cn}$) &	\ub{92.00} &	89.78 &	54.89 &	75.12 &	39.04 &	\ux{70.17} \\
\quad Ours ($\hat{T}^{*}_{csgcn}$) &	91.76 &	\ub{90.59} &	50.95 &	74.52 &	35.59 &	68.68 \\ \midrule
\textbf{Transfer} \\
\quad $T = 1$ &	90.17 &	\ux{96.31} &	90.55 &	\ux{90.19} &	\ub{65.85} &	86.61 \\
\quad $T = \sqrt{M}$ &	86.04 &	34.79 &	16.44 &	62.47 &	33.67 &	46.68 \\
\quad Insert LN \cite{hase} &	\ux{92.13} &	96.16 &	\ux{90.76} &	89.51 &	64.52 &	\ux{86.62} \\
\quad Ours ($\hat{T}^{*}$) &	92.03 &	93.01 &	76.73 &	89.34 &	65.27 &	83.28 \\
\quad Ours ($\hat{T}^{*}_{csg}$) &	91.39 &	95.47 &	83.57 &	90.11 &	\ux{65.59} &	85.22 \\
\quad Ours ($\hat{T}^{*}_{cn}$) &	\ub{92.71} &	95.81 &	88.08 &	90.05 &	64.95 &	86.32 \\
\quad Ours ($\hat{T}^{*}_{csgcn}$) &	91.69 &	\ub{96.39} &	\ub{91.33} &	\ub{90.24} &	65.05 &	\ub{86.94} \\
        \bottomrule
    \end{tabular}
\end{table}

\section{Conclusion}
In this study, we investigated the optimization of the temperature parameter in the commonly used softmax cross-entropy loss for classification problems using deep learning. It has been suggested that there is an important relationship between the number of dimensions $M$ of the encoder's feature map and the temperature $T$, and we aimed to theoretically verify this relationship and experimentally determine the optimal temperature $T^*$. Theoretical verification led to the hypothesis that standardizing the feature maps using BN enables robust calculation of the optimal temperature $T^*$ across different models and datasets. The experimental results support this hypothesis, demonstrating that BN insertion can lead to optimal performance, whereas using $T = 1$ often results in performance degradation. 

Based on a comprehensive evaluation, we propose a method for estimating \(T^*\) through optimization, along with a correction approach based on task difficulty (denoted by CSG) and the number of classes (denoted by CN), thereby yielding an optimal temperature \(\hat{T}^*\) that leads to performance improvement.
Ultimately, we derived the formula
\[
\hat{T}^*_{csgcn} = \operatorname{clip}\bigl(0.3192 \sqrt{M} + 20.74 + 3.746 \log(csg) - 7.380 \log(cn), \epsilon, 512\bigr).
\]
Evaluations across various model architectures and image classification datasets demonstrated that the proposed method consistently achieves high accuracy in most cases.

This study has several limitations. First, it focuses solely on image classification datasets. 
The related work~\cite{hase} has demonstrated that similar phenomena occur in one-dimensional waveform activity recognition. This suggests that the proposed method may also be effective in other tasks employing the softmax cross-entropy loss, such as speech and document classification. Second, there is room for improvement in the correction formula from ($\hat{T}^*$ to $\hat{T}^*_{csgcn}$). The temperature determination coefficients were optimized based on CIFAR and Tiny ImageNet experiments. Therefore, diversifying measurement points and environments may lead to more generalized temperature determination coefficients. Currently, the temperature determination coefficients are optimized using data from up to 200 classes in the Tiny ImageNet dataset. Consequently, their effectiveness cannot be assured for scenarios involving a larger number of classes or substantially larger datasets. Furthermore, the correction for task difficulty was performed using the CSG, which involves some ambiguity and uncertainty. Therefore, we plan to further explore dynamic adjustments to task difficulty based on interclass similarity during training. Third, the temperature $T$ was kept constant in this study. Given the characteristics of $T$, it may be preferable for $T$ to be lower during the early stages of training and increase as the training progresses. We plan to investigate the implications of a dynamically changing $T$ during training in the future work.

\begin{appendices}
\section{Inserting BN}\label{sec:a1}
Normalization layers, such as BN \cite{bn} and LN \cite{ba2016layer}, aim to normalize feature maps. These layers mitigate the distribution shift between layers, referred to as the internal covariate shift. These normalization layers have been introduced by recent architectures, such as ViT \cite{dosovitskiy2020vit} and the Swin transformer (Swin-T) \cite{swint}.
However, it is uncommon to insert normalization directly before the output layer, as proposed in this study. Referring to official PyTorch implementations\footnote{PyTorch: Models and pre-trained weights [\url{https://pytorch.org/vision/main/models.html}]}, many models, such as ResNet \cite{resnet}, RegNet \cite{Radosavovic_2020_CVPR}, and EfficientNet \cite{pmlr-v97-tan19a}, typically connect the output layer after an encoder. Among the models we examined, ConvNeXt \cite{Liu_2022_CVPR}, ViT and Swin-T connect the output layer following the sequence: encoder $\rightarrow$ GAP $\rightarrow$ LN.

Based on our hypothesis described in the previous section \ref{sec:hypothesis}, we employ BN as the normalization layer, as it normalizes features for the number of samples. Inserting a normalization layer immediately before the output layer has not traditionally been a common practice and has only been adopted in certain specific model architectures mentioned above. While the insertion of BN is pointed out to have a stabilizing effect on temperature determination coefficients, there are concerns that it may degrade the original performance, necessitating verification (we discuss in Section \ref{sec:bnexp}).

\section{Additional Theoretical Justification of the Temperature Coefficients}
\update{This appendix details the derivation of the two task–dependent correction terms---CSG correction and class‑number (CN) correction---that appear in Eq.~\eqref{eq:final}.}

\subsection{CSG Correction Term}
\label{app:csg_corr_en}
\update{With BN inserted immediately before the output layer, the logit variance is given by Eq.~\eqref{eq:var5}, reproduced here for convenience:
\[
  \mathbb{V}\!\bigl[\hat y_j\bigr]
    = \sum_{k=1}^{M} w_{jk}^{2}
      + \sum_{k \neq l} w_{jk} w_{jl}\,
        \mathbb{E}[\,z_k z_l\,].
\tag{A.2}\label{eq:resid_var_en}
\]
The first summation depends only on the architecture, whereas the second captures the residual inter‑feature correlations. Empirically, these correlations increase with task difficulty, and can be approximated by
\[
  \sum_{k \neq l}
     w_{jk} w_{jl}\,\mathbb{E}[z_k z_l]
  \;\propto\;
  \log\!\bigl(\mathrm{CSG}\bigr).
\tag{A.3}\label{eq:csg_prop_en}
\]
Because a larger logit variance flattens the soft‑max distribution---effectively acting as a higher temperature---the temperature must be lowered in accordance with task difficulty to counter this effect. We therefore adopt CSG as a proxy measure of task difficulty.}

\subsection{Class-Number (CN) Correction Term}
\label{app:cn_corr_en}
\begin{figure}[h]
\centering
 \includegraphics[width=0.5\textwidth]
      {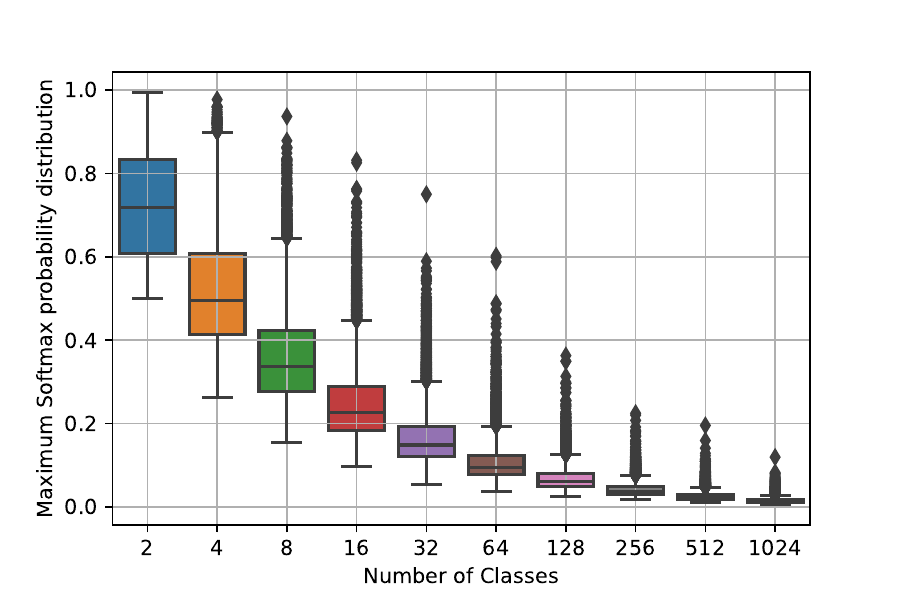}
 \caption{Distribution of the maximum softmax probability for varying class counts.}
 \label{fig:top_prob_sim}
\end{figure}

\update{Let \(\mathbf y=(y_{1},\dots,y_{C})\) be row--wise i.i.d.\ logits and \(p_{i}=e^{y_{i}/T}\bigl/\sum_{k=1}^{C}e^{y_{k}/T}\) the corresponding softmax probabilities. Because \(\sum_{i=1}^{C}p_{i}=1\) and the distribution of \(\mathbf y\) is exchangeable, the unconditional expectation of any single class probability is $\mathbb{E}[p_{i}] \;=\; \frac{1}{C}$. This equation shows that the baseline confidence allocated to each class decays inversely with the number of classes \(C\). Consequently, the maximum softmax probability in a sample \(\max_{j}p_{j}\) also decreases monotonically with \(C\), a tendency confirmed by the corrected simulation in Fig.~\ref{fig:top_prob_sim}.}

\update{For the cross-entropy loss \(L=-\log p_{y^{\ast}}\) (\(y^{\ast}\) is the ground-truth label), the gradient with respect to a logit \(z_{i}\) is
\[
   \frac{\partial L}{\partial z_{i}}
   \;=\;
   \frac{1}{T}\bigl(p_{i}-\mathbbm{1}_{\{i=y^{\ast}\}}\bigr).
\]
If \(C\) grows while the logit scale remains unchanged, \(p_{y^{\ast}}\) converges to \(1/C\) and \(\partial L/\partial z_{i}\) shrinks approximately as \(1/T\,C\). The optimisation thus suffers from vanishing gradients. One remedy is to reduce \(T\) (i.e.\ sharpen the softmax) so that the
effective gradient magnitude stays within a favourable range. Hence, \(T^{*}\) must decrease in expectation with \(C\).}

\update{Because the effect described above is systematic and orthogonal to the feature-dimension effect (first term), we embed \(C\) in the model through the logarithmic correction term \(\delta\log(\text{cn})\) in Eq.~\eqref{eq:final}. In Fig.~\ref{fig:top_prob_sim}, we show the distribution of the maximum softmax probability \(\max_{j}p_{j}\) as a function of the number of classes \(C\).  For clarity, the horizontal axis is plotted on a \(\log_{2}\) scale so that the powers of two (2, 4, 8, $\cdots$, 1024) appear evenly spaced.  This choice is purely empirical and serves to improve readability; it is not derived from any grid‑search procedure.  The plot confirms that \(\max_{j}p_{j}\) decays monotonically toward zero as \(C\) increases, although the decay is not strictly linear in \(\log C\). Optimising the coefficient \(\delta\) across diverse datasets yields the globally valid estimate reported in Table~\ref{table:coeff}.}

\end{appendices}

\section*{Acknowledgement}
This work was supported in part by the Japan Society for the Promotion of Science (JSPS) KAKENHI Grant-in-Aid for Scientific Research (C) under Grants 23K11164.

\textbf{Availability of data and material}
We used publicly available datasets CIFAR10/100 \cite{cifar10}, STL10 \cite{stl10}, Tiny ImageNet (Tiny IN) \cite{imagenet}, German Traffic Sign Recognition Benchmark (GTS)\cite{gtsrb}, EuroSAT (SAT)\cite{helber2019eurosat}, ISIC Challenge 2019 (ISIC)\cite{Tschandl2018, Codella2018, Hernandez2024}, Caltech101 (CT)\cite{caltech}, Flowers102 (FLW)\cite{flowers}, Oxford-IIIT Pet (PET)\cite{pet}, Describable Textures Dataset (DTD)\cite{dtd}, CUB-2011-200 dataset (CUB)\cite{WahCUB_200_2011} for our experiments. All datasets are available for each web site.

\textbf{Competing interests}
The authors declare that they have no conflicts of interest.

\bibliographystyle{unsrt}  
\bibliography{sn-bibliography}

\end{document}